\documentclass[letterpaper]{article} 
\usepackage{aaai23}  
\usepackage{times}  
\usepackage{helvet}  
\usepackage{courier}  
\usepackage[hyphens]{url}  
\usepackage{graphicx} 

\usepackage{color}

\usepackage{natbib}  
\usepackage{caption} 
\usepackage{algorithm}
\usepackage{algorithmic}

\newcommand{\ie}{\emph{i.e.,}~}

\newcommand{\blue}[1]{\textcolor{black}{#1}}

\usepackage{newfloat}
\usepackage{listings}
\DeclareCaptionStyle{ruled}{labelfont=normalfont,labelsep=colon,strut=off} 
\floatstyle{ruled}
\newfloat{listing}{tb}{lst}{}
\floatname{listing}{Listing}

\usepackage{booktabs}
\usepackage{multirow}
\usepackage{amssymb}
\usepackage{subfigure}
\usepackage{amsmath}
 \usepackage[backref]{hyperref} 
 \hypersetup{
hidelinks,
}

\title{Hierarchical Contrast for Unsupervised Skeleton-based Action \\ Representation Learning}

\author{
    Jianfeng Dong\textsuperscript{\rm 1,2}, 
    Shengkai Sun\textsuperscript{\rm 1}, 
    Zhonglin Liu\textsuperscript{\rm 1}, 
    Shujie Chen\textsuperscript{\rm 1,2}\thanks{Corresponding author}, 
    Baolong Liu\textsuperscript{\rm 1,2}, 
    Xun Wang\textsuperscript{\rm 1,2} 
}
\affiliations{
    \textsuperscript{\rm 1}College of Computer Science and Technology, Zhejiang Gongshang University, China\\
    \textsuperscript{\rm 2}Zhejiang Key Lab of  E-Commerce, China \\
}

\usepackage{bibentry}

\begin{document}

\maketitle

\begin{abstract}
This paper targets unsupervised skeleton-based action representation learning and proposes a new Hierarchical Contrast (HiCo) framework. Different from the existing contrastive-based solutions that typically represent an input skeleton sequence into instance-level features and perform contrast holistically, our proposed HiCo represents the input into multiple-level features and performs contrast in a hierarchical manner. Specifically, given a human skeleton sequence, we represent it into multiple feature vectors of different granularities from both temporal and spatial domains via sequence-to-sequence (S2S) encoders and unified downsampling modules. Besides, the hierarchical contrast is conducted in terms of four levels: instance level, domain level, clip level, and part level. Moreover, HiCo is orthogonal to the S2S encoder, which allows us to flexibly embrace state-of-the-art S2S encoders. Extensive experiments on four datasets, \ie NTU-60, NTU-120, PKU-MMD I and II, show that HiCo \blue{achieves} a new state-of-the-art for unsupervised skeleton-based action representation learning in two downstream tasks including action recognition and retrieval, and its learned action representation is of good transferability. Besides, we also show that our framework is effective for semi-supervised skeleton-based action recognition. \blue{Our code is available at \url{https://github.com/HuiGuanLab/HiCo}.}
\end{abstract}

\section{Introduction}\label{sec:introduction}
\begin{figure}[tb!]
\subfigure[A typical contrastive learning framework]{
\includegraphics[width=0.99\columnwidth]{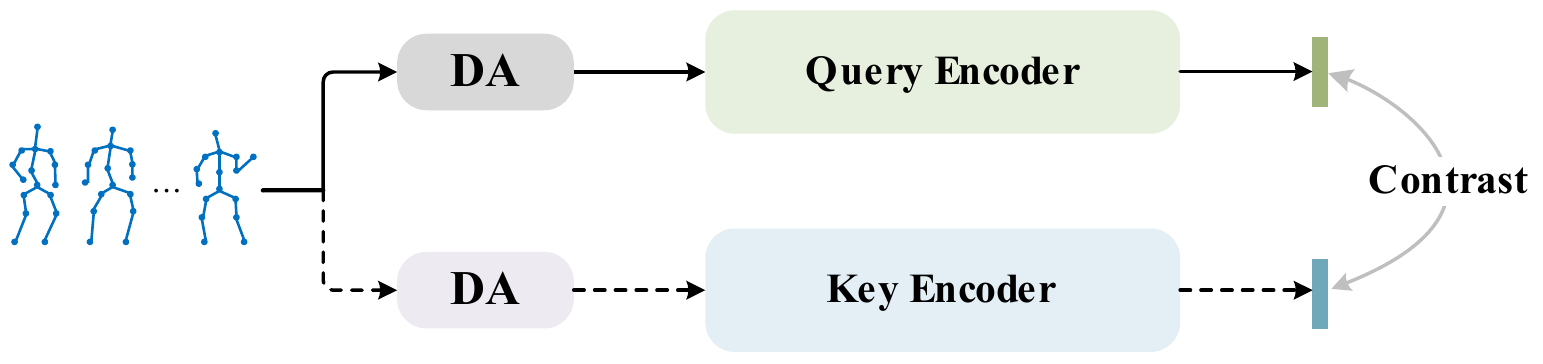}
}
\subfigure[Our proposed HiCo framework]{
\includegraphics[width=0.99\columnwidth]{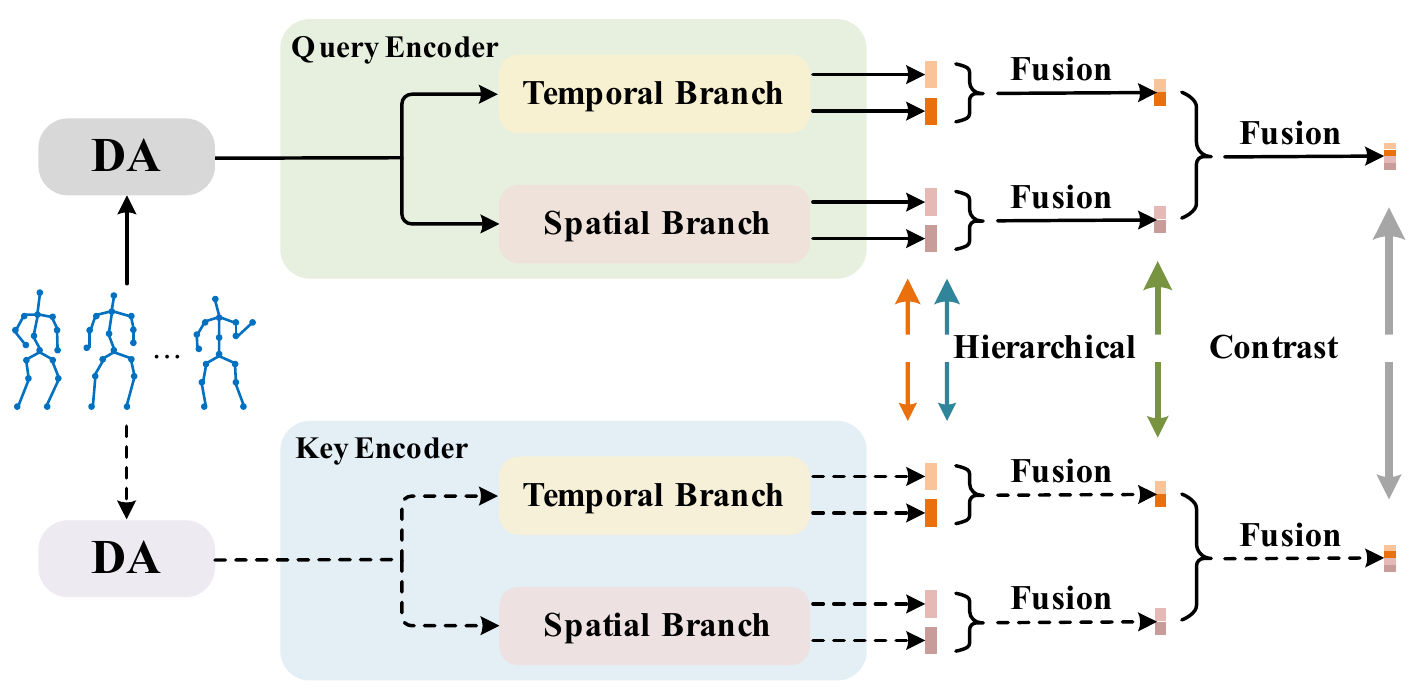}
}
\caption{Frameworks for unsupervised SARL. (a) A typical contrastive learning framework that represents skeleton instances at instance-level features, and performs contrast holistically. (b) Our proposed HiCo represents them at multi-level features, and performs contrast hierarchically. DA indicates data augmentation.
}\label{fig:replace_experiment}
\end{figure}

Human action recognition has a broad application in fields of human-computer interaction, intelligent surveillance, video content analysis, game control, etc~\cite{gao2019know,li2019w2vv++,dong2021dual}. In recent years, 3D skeleton-based action recognition has made remarkable progress with deep learning networks 
\cite{zhang2012microsoft,liu2019ntu,wang2020learning}. 
However, how to learn a more discriminative representation of skeletons is still an open question for skeleton-based action recognition.
To conquer this question, the majority of early works \cite{ni2011rgbd, wang2012mining, vemulapalli2016rolling, zhang2017view, liu2019ntu,cheng2020skeleton} trained a network in a fully-supervised manner, which required massive annotated 3D skeleton data and hence is expensive and time-consuming. 
Recently, we observe a new trend of proposing unsupervised Skeleton-based Action Representation Learning (SARL) method to liberate the workload of annotation~\cite{zheng2018unsupervised, su2020predict, lin2020ms2l}. In this paper, we also focus on unsupervised SARL.

Existing efforts on unsupervised SARL can be roughly categorized into three groups: encoder-decoder methods~\cite{zheng2018unsupervised,nie2020unsupervised}, contrastive-learning methods~\cite{rao2021augmented,thoker2021skeleton,guo2022contrastive}, and hybrid methods~\cite{su2021self,chen2022hierarchically}.
The encoder-decoder methods first encode the input skeleton sequence into latent features, then decode the latent features under the guidance of various hand-crafted pretext tasks, such as skeleton reconstruction~\cite{zheng2018unsupervised}, skeleton colorization prediction~\cite{yang2021skeleton}, and skeleton displacement prediction~\cite{kim2022global}.
The contrastive-learning methods usually augment an input skeleton sequence into two augmented instances, and train an encoder to make the instances of the same skeleton have more similar representations than that of different skeletons.
The hybrid methods integrate the ideas of both encoder-encoder and contrastive-learning.
Among them, the contrastive-learning methods are dominated in recent years due to their simple mechanism and superior performance~\cite{li20213d,guo2022contrastive}.

The first contrastive-learning method for unsupervised SARL was proposed by~\cite{rao2021augmented}, which adapted MoCo~\cite{he2020momentum} that was originally designed for unsupervised image representation learning to unsupervised skeleton representation learning.
After that, a number of improved contrastive-learning works were proposed by modeling the uncertainty of skeletons~\cite{su2021modeling}, exploring more positive pairs~\cite{li20213d}, and utilizing superior skeleton-specific augmentations~\cite{thoker2021skeleton}.
As illustrated in Figure \ref{fig:replace_experiment}(a), these contrastive-learning methods usually first represent skeleton sequences into instance-\blue{level} features, and then perform instance-level contrast \textit{holistically}. 
Although the contrastive-learning methods have shown much better performance, such holistic instance-level contrast may be suboptimal considering human skeletons naturally have \textit{hierarchical} structures.
A skeleton sequence can be commonly regarded as a list of whole skeletons (frames) \blue{temporally}, or a list of skeleton joints \blue{spatially}. Moreover, either frames or joints are the fundamental elements in the temporal or spatial domain, which can be built as larger-granularity elements such as frame clips or body parts.

Inspired by the hierarchical nature of human skeletons, we propose a new framework Hierarchical Contrast (HiCo) for unsupervised SARL.
HiCo has a hierarchical encoder network to encode skeletons into part-level, clip-level, domain-level, and instance-level representations, and perform multiple-level contrast hierarchically, as shown in Figure \ref{fig:replace_experiment}(b).
Specifically, the hierarchical encoder network has two branches corresponding to the temporal domain and the spatial domain respectively.
For the temporal branch, it encodes a skeleton sequence into multiple clip-level features of different temporal granularities by extracting features from clips of various lengths. 
Similarly, the spatial branch extracts features from human parts of various sizes to obtain part-level features.
Furthermore, the clip-level features and part-level features derived from temporal and spatial domains are progressively used to compose the domain-level representation and the instance-level representation.
Additionally, given the multiple-level representations, a hierarchical contrast is conducted in terms of four levels.
Our hypothesis is that such hierarchical contrast is consistent with the hierarchical nature of human skeletons and provides more supervision which is crucial for unsupervised learning.
In summary, this paper makes the following contributions.
\begin{itemize}
\item We propose a hierarchical encoder network that represents skeleton sequences into multiple feature vectors of different granularities from both temporal and spatial domains via sequence-to-sequence (S2S) encoders and unified downsampling
modules. Moreover, our framework is orthogonal to the S2S encoder, which allows us to flexibly embrace state-of-the-art S2S encoders.

\item Based on the multi-level representations of human skeletons, we propose a new hierarchical contrastive loss for unsupervised learning, which is more effective than the plain contrastive loss over instance-level representation in the context of unsupervised SARL.

\item Extensive experiments on four datasets show that our HiCo \blue{achieves} a new state-of-the-art for unsupervised SARL in two downstream tasks, and its learned representation is of good transferability. Besides, we also show that our method is also effective for semi-supervised skeleton-based action recognition.
\end{itemize}

\section{Related Work} \label{sec:rel-work}
\subsection{Unsupervised SARL} 
Unsupervised methods has achieved increasing attention in various tasks~\cite{yang2018semantic,deng2019unsupervised,gao2020unsupervised,liu2022unsupervised} due to its good characteristic of training without human annotations. Existing works on unsupervised SARL can be roughly categorized into three groups: 
encoder-decoder methods, contrastive-learning methods, and hybrid methods.

\subsubsection{Encoder-decoder methods.}
The early works mainly adopt the encoder-decoder paradigm, which typically encodes the input skeleton sequence into latent features, then decodes the latent features under the guidance of various hand-crafted pretext tasks. The key to this paradigm is how to design a pretext task for unsupervised SARL.
Skeleton reconstruction is popular as the pretext task. For instance, \cite{zheng2018unsupervised} learned to reconstruct the whole skeleton from the corrupted skeleton, \cite{kundu2019unsupervised, su2020predict} reconstructed from the whole skeleton via auto-encoders. \cite{nie2020unsupervised} first disentangled the skeleton into the pose-dependent and view-dependent features, then reconstructed skeletons from the disentangled features.
Recently, a number of novel pretext tasks have been proposed. \cite{su2021self} colorized each joint of human skeletons according to its temporal and spatial orders, and a skeleton colorization prediction pretext task was employed. \cite{cheng2021hierarchical} devised a motion prediction pretext task that predicted the motion between adjacent frames. 
In \cite{kim2022global}, a multi-interval pose displacement prediction pretext task was proposed, which aimed to estimate the whole-body and joint motions at various time scales.

\begin{figure*}[tb!]
\centering\includegraphics[width=1.99\columnwidth]{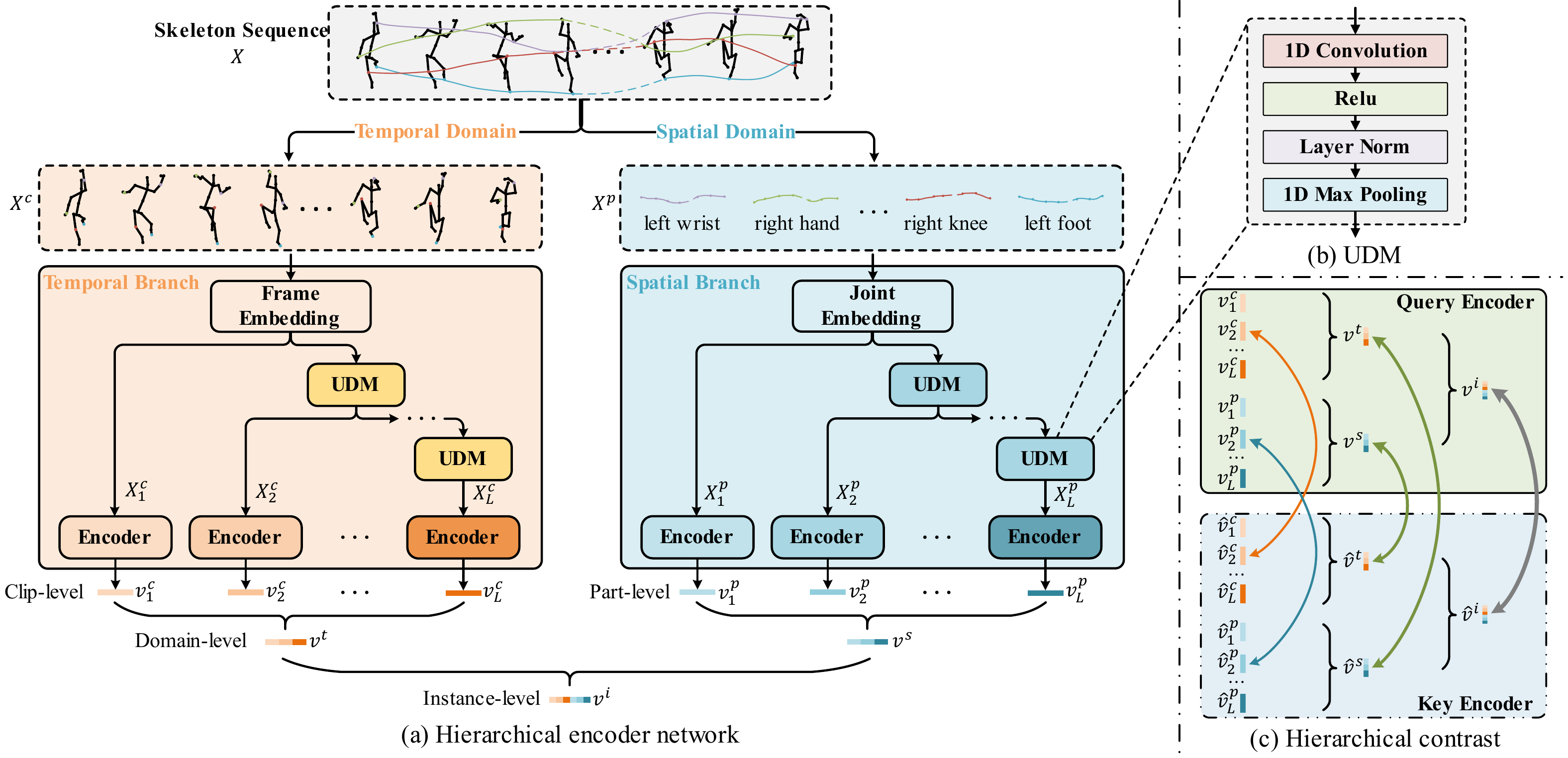}
\caption{
The illustration of our HiCo framework. It consists of (a) a hierarchical encoder network using S2S encoders with our designed (b) unified downsampling modules (UDM), which encodes a skeleton sequence at multiple granularities of both temporal and spatial domains.
(c) a hierarchical contrast that conducts contrastive learning in terms of four levels jointly.
Note that encoders are shared in the same branch.}\label{fig:framework}
\end{figure*}

\subsubsection{Contrastive-learning methods.}
\cite{rao2021augmented} was the first to introduce contrastive learning for unsupervised SARL.
In \cite{rao2021augmented}, a skeleton sequence was first transformed into two instances by random augmentation strategies, which were further fed into a query encoder and a key encoder respectively to obtain instance-level features. The contrast was conducted between the outputs of both the key and query encoders, encouraging augmentations of the same skeleton (positive pairs) to have more similar representations than that of different skeletons (negative pairs).
Besides the positive pairs generated by random augmentation, \cite{li20213d} exploited the multi-view knowledge of skeleton data for mining more positive pairs.
\cite{thoker2021skeleton} proposed skeleton-specific spatial and temporal augmentations to enhance the spatio-temporal invariances of the learned representation. Moreover, three encoders of different architectures were utilized to conduct cross-architecture contrasts.
In \cite{su2021modeling}, they represented skeleton sequences into a probabilistic embedding space, and the similarity was measured as the distance between probabilistic distributions.

\subsubsection{Hybrid methods.}
Hybrid methods combine the idea of the encoder-decoder and contrastive-learning together~\cite{su2021self,xu2021prototypical}.
A typical method was proposed by \cite{lin2020ms2l}, where besides the contrastive learning between skeletons, a motion prediction and a jagsaw puzzle recognition pretext tasks were jointly employed. Different from \cite{lin2020ms2l} that added pretext tasks on the final learned features, \cite{chen2022hierarchically} employed pretext tasks on the outputs of middle layers.

\subsection{Hierarchical Modeling on Skeletons}
There are also some previous works that modeled human skeletons hierarchically. For instance, \cite{du2015hierarchical,cheng2021hierarchical,wang2020learning} modeled the spatial hierarchical structure with hand-crafted designs, but they did not consider the temporal hierarchical structure.
\cite{chen2022hierarchically} exploited the temporal hierarchical clues from the frame, clip, and video levels.
By contrast, we jointly model the temporal and spatial hierarchical structure, which can be readily implemented using our proposed more unified framework. Besides, different from the previous models that present skeletons at the instance level, we present skeletons at the instance level, domain level, clip level, and part level.

\section{Method} \label{sec:method}
In this paper, we propose a hierarchical contrast framework for unsupervised SARL.
As shown in Figure \ref{fig:framework}, our framework contains two main components: 
1) a hierarchical encoder network that encodes a skeleton sequence at multiple granularities of both temporal and spatial domains, deriving the hierarchical multi-level representations.
2) a hierarchical contrast for unsupervised learning, which conducts contrastive learning in terms of four levels jointly.
In what follows, we first depict the hierarchical encoder network, followed by the description of the hierarchical contrast.

\subsection{Hierarchical Encoder Network}

The hierarchical encoder network consists of a temporal branch and a spatial branch, which encodes the input skeleton sequence from the temporal domain and the spatial domain, respectively. The temporal branch encodes a skeleton sequence at multiple \textit{temporal} granularities, resulting in a clip-level representation. Similarly, the spatial branch encodes it at multiple \textit{spatial} granularities, resulting in a part-level representation.
Both the clip-level representation and part-level representation are progressively used to assemble the domain-level and instance-level representations.

\subsubsection{Clip-level Representation.} 

Given a skeleton sequence $X\in\mathbb{R}^{T\times J\times 3}$ that has $T$ frames with $J$ joints, and each joint consists of three spatial coordinates, we first reshape it in the time-majored domain as a list of frames $X^c=\{x_i^c\}_{i=1}^{T} \in \mathbb{R}^{T\times 3J}$, where $x_i^c$ denotes the $i$-th frame that is essentially a whole human skeleton.
Besides, a frame embedding is employed to project the input into a $C$-dimensional dense feature space by two fully connected layers. Formally, a projected feature from $x_i^c$ is obtained as:
\begin{equation}
\widetilde{x_i^c} = W_2(\sigma(W_1 x_i^c + b_1))+b_2,
\end{equation}
where $W_1 \in \mathbb{R}^{C\times 3J}$ and  $W_2 \in \mathbb{R}^{C\times C}$ are transformation matrix, $b_1 \in \mathbb{R}^{C}$ and $b_2 \in \mathbb{R}^{C}$ indicate the bias vectors, and $\sigma$ is the ReLU activation function. All projected features are jointly denoted as $X^c_1=\{\widetilde{x_i^c}\}_{i=1}^{T} \in \mathbb{R}^{T\times C}$. For ease of reference, we name these features as the initial frame representation.
As the initial frame features are represented independently, they naturally lack temporal dependencies that are crucial for action representation. Therefore, we propose to model their temporal dependency with multiple temporal granularities. We first generate clips of different temporal granularities, and then utilize S2S encoders to model their temporal dependencies.

\textbf{\textit{Clip Construction}}. 
In order to generate clips of different temporal granularities, we propose a Unified Donwsampling Module (UDM), which sequentially merges multiple consecutive frame/clip features to obtain more coarse-grained clips.
The structure of UDM is illustrated in Figure \ref{fig:framework}(b), which can be formally defined as:
\begin{equation}
\text{UDM}(\cdot)=\text{MaxPool1D}(\text{LN}(\sigma(\text{Conv1D}(\cdot)))),
\end{equation}
where Conv1D indicates a 1D convolution layer with a kernel size of 5 and stride size of 1, LN denotes the Layer Norm, and MaxPool1D is the 1D max pooling with kernel size 2.
Here the 1D convolution layer is used to collect nearby contextual information and the max pooling is used for information aggregation.
Note that UDM can be stacked in series, which leads to more coarse-grained clips. Additionally, we regard frames as single-frame clips at the granularity of 1 for ease of description. With the help of UDM, given clips at the granularity of $n$, clips of larger granularity can be readily obtained as:
\begin{equation}
X^c_{n+1}=\text{\blue{UDM}}(X^c_{n}). 
\end{equation}
By jointly employing UDM $L\mathit{-}1$ times in series, we are able to obtain clips of different temporal granularities $\{X^c_{1}, X^c_{2},\dots, X^c_{L} \}$.

\textbf{\textit{Temporal Dependency Modeling.}} 
For each granularity of clips, we further model the temporal dependency by an S2S encoder. 
Concretely, given clips $X^c_{n}$, we feed it into an S2S encoder, such as GRU, followed by a temporal max pooling layer (TMP) for feature aggregation, and the clip-level feature $v^c_n$ at the granularity of $n$ is obtained as:
\begin{equation}
v^c_n=\text{TMP}(\text{S2S}(X^c_n)).
\end{equation}
Here, the outputs from all time steps of the S2S encoder are used.
Accordingly, given $\{X^c_{1}, X^c_{2},\dots, X^c_{L} \}$ of $L$ granularities, we are able to obtain the final clip-level representation $V^c$ containing $L$ feature vectors as:
\begin{equation}
V^c = \{v_1^c, v_2^c,\dots, v_L^c\}.
\end{equation}
Note that our network is orthogonal to the S2S encoder, and any S2S encoders can be employed theoretically. In our implementation, we try GRU, LSTM, and Transformer.

\subsubsection{Part-level Representation.}
Obtaining the part-level representation is similar to obtaining the clip-level representation, so we mainly specify choices that are unique at the part-level.
Given a skeleton sequence $X\in\mathbb{R}^{T\times J\times 3}$, here we reshape it in the space-majored domain as a list of joints $X^p=\{x_i^p\}_{i\in J}\in \mathbb{R}^{J\times 3T}$, where $x_i^p$ denotes the $i$-th joint that is essentially a joint trajectory along the time.
Similarly, a joint embedding is employed to obtain the initial joint representation $X^p_1 \in \mathbb{R}^{J\times C}$.

Additionally, we generate human parts of different spatial granularities by UDM. The UDM here is employed to merge nearby joints/parts to obtain parts of larger sizes. Similarly, by jointly employing UDM $L\mathit{-}1$ times in series, we are able to obtain parts of different spatial granularities $\{X^p_{1}, X^p_{2},\dots, X^p_{L} \}$. Furthermore, for each granularity, an S2S encoder with a max pooling layer is used to capture the spatial dependency.
Consequently, we are able to obtain the final part-level representation containing clues of multiple spatial granularities as $V^p = \{v_1^p, v_2^p,\dots, v_L^p\}$.

\subsubsection{Domain-level and Instance-level Representation.} 
Thus far we have a clip-level representation of multiple temporal granularities $V^t$ and a part-level representation of multiple spatial granularities  $V^s$. We further combine them progressively to get the domain-level representation and the instance-level representation.

For the domain-level representation, we fuse all $L$ feature vectors of clip-level representation to derive the temporal-domain representation $v^t$, and fuse all the corresponding feature vectors of part-level representation to derive the spatial-domain representation $v^s$:
\begin{equation}
\begin{aligned}
v^t = F(V^t) = F(v_1^c, v_2^c,\dots, v_L^c), \\
v^s = F(V^p) = F(v_1^p, v_2^p,\dots, v_L^p),
\end{aligned}
\end{equation}
where $F(\cdot)$ denotes the fusion operator over multiple feature vectors. The concatenation is used in our implementation.

Moreover, the temporal-domain and spatial-domain representations are jointly combined into the instance-level representation $v^i$:
\begin{equation}
v^i = F(v^t, v^s).
\end{equation}
It is worth pointing out that though multiple-level representations have been derived, only the instance-level representation is used for the final skeleton sequence representation. Other representations are used for hierarchical contrast, which will be described in the following section.

\subsection{Hierarchical Contrast}
Different from the previous contrastive-learning methods that conduct contrast only at instance-level features, we propose a hierarchical contrast that conducts contrast at instance-level, domain-level, clip-level, and part-level features.
For the contrastive way, we adopt the MoCo manner \cite{he2020momentum} that a query encoder and a key encoder are jointly used with the dynamic dictionary queue and moving-averaged update mechanism.
Besides, following \cite{chen2020simple}, we utilize a two-layer multilayer perceptron for feature projection before contrast.
For concise descriptions, we use the same symbol of the projected features as the original ones. We utilize $\hat{}$ to indicate the features derived from the key encoder, and $m$ to denote the feature from the queue.

\subsubsection{Instance-level Contrast.} 
Following~\cite{rao2021augmented}, we use the noise contrastive estimation loss InfoNCE \cite{van2018representation} for contrast. At the instance level, the contrastive loss is computed as: 
\begin{small}
\begin{equation}
\mathcal{L}_{Instance}=-\log\frac{\exp(v^i \cdot \hat{v^i}/\tau)}
 {\exp(v^i \cdot \hat{v^i}/\tau)+\sum_{m^{i}_j\in M^{i}}^{} \exp(v^i \cdot m^i_j/\tau)},
\end{equation}
\end{small}
where $\tau$ is the temperature hyper-parameter, $m^{i}_j$ denotes $j$-th negative samples from the first-in-first-out queue $M^{i}$ of previous projected instance-level features.

\subsubsection{Domain-level Contrast.} 
At the domain level, we expect a skeleton sequence have similar temporal-domain and spatial-domain representations.
Our intuition is that the temporal-domain and spatial-domain representations are descriptions from different perspectives of the same skeleton sequence, so they essentially should be similar in high-level semantics. Therefore, we add a contrastive loss over two cross-domain features $v^t$ and $v^s$, which encourages the network to pull the temporal-domain and spatial-domain features of the same skeleton sequence while pushing away that of the different skeleton sequences:

\begin{small}
\begin{equation}
\begin{aligned}
\mathcal{L}_{domain}=&-\log\frac{\exp(v^t \cdot \hat{v^s}/\tau)}
 {\exp(v^t \cdot \hat{v^s}/\tau)+\sum_{m^{s}_j\in M^{s}}^{} \exp(v^t \cdot m^{s}_j/\tau)}\\
 &-\log\frac{\exp(v^s \cdot \hat{v^t}/\tau)}
 {\exp(v^s \cdot \hat{v^t}/\tau)+\sum_{m^{t}_j\in M^{t}}^{} \exp(v^s \cdot m^{t}_j/\tau)}.
\end{aligned}
\end{equation}
\end{small}

\subsubsection{Clip-level Contrast.} 
Recall that clip-level representation has multiple-feature vectors of different temporal granularities. Although they are at different granularities, 
they also should be similar in high-level semantics. Therefore, we consider the clip-level features of different granularities of the same input instance positive. For ease of implementation, the clip-level feature at the granularity of 1 is regarded as the anchor sample, which is positive with other granularities of the same input. 
The contrastive loss at the clip-level is obtained by InfoNCE with more positive pairs:
\begin{small}
\begin{equation}
\mathcal{L}_{clip}=-\log\frac{\sum_{l=1}^{L}\exp(v^c_1 \cdot \hat{v^c_l}/\tau)}
 {\sum_{l=1}^{L}\exp(v^c_1 \cdot \hat{v^c_l}/\tau)+\sum_{m^{c}_j\in M^{c}}^{} \exp(v^c_1 \cdot m^{c}_j/\tau)},
\end{equation}
\end{small}
where $M^{c}$ denotes the queue containing previous projected clip-level features of all granularities. 

\subsubsection{Part-level Contrast.} Similar to the clip-level contrast, the contrastive loss at
the part-level is computed on part-level representation, which is defined as:
\begin{small}
\begin{equation}
\mathcal{L}_{part}=-\log\frac{\sum_{l=1}^{L}\exp(v^p_1 \cdot \hat{v^p_l}/\tau)}
 {\sum_{l=1}^{L}\exp(v^p_1 \cdot \hat{v^p_l}/\tau)+\sum_{m^{p}_j\in M^{p}}^{} \exp(v^p_1 \cdot m^{p}_j/\tau)}.
\end{equation}
\end{small}

Finally, we train our network by minimizing the sum of the above losses. The total loss is computed as:
\begin{equation*}
\mathcal{L}_{total}= \mathcal{L}_{instance}+\mathcal{L}_{domain}+\mathcal{L}_{clip}+\mathcal{L}_{part}.
\end{equation*}

\section{Experiments} \label{sec:eval}
\subsection{Experimental Setup} \label{ssec:exp-dataset}
\subsubsection{Datasets} \label{ssec:exp-dataset}
Experiments are conducted on four popular skeleton-based action datasets, \ie NTU-60 \cite{shahroudy2016ntu}, NTU-120 \cite{liu2019ntu}, PKU-MMD I, and PKU-MMD II \cite{liu2020benchmark}.

\subsubsection{Evaluation Metrics}
On NTU-60 and NTU-120, we follow two standard evaluation protocols: cross-subject (x-sub), and cross-view (x-view). Following \cite{lin2020ms2l}, we report the x-sub results on PKU-MMD I and II.
The top-1 accuracy is used for performance evaluation.

\begin{table*} [tb!]
\renewcommand{\arraystretch}{1.2}
\caption{ Comparisons to the state-of-the-art methods for skeleton-based action recognition downstream task on NTU-60, NTU-120, PKU-MMD I and II. 
Compared methods are sorted in ascending order in terms of their x-sub performance on NTU-60.
Symbol asterisk (*) indicates numbers obtained by fusing three-stream models of joint, bone and motion views, and the others only utilize the joint view of skeletons as input.
} 
\label{tab:sota-unsupervised}
\centering \scalebox{0.8}{
\begin{tabular}{@{}l*{9}c @{}}
\toprule
\multirow{2}{*}{\textbf{Method}}   &
\multirow{2}{*}{\textbf{Type}}  &
\multirow{2}{*}{\textbf{Encoder}}  &
\multicolumn{2}{c}{\textbf{NTU-60}} &
\multicolumn{2}{c}{\textbf{NTU-120}} &
\multicolumn{1}{c}{\textbf{PKU-MMD I}} &
\multicolumn{1}{c}{\textbf{PKU-MMD II}} 
\\
\cmidrule(r){4-5} \cmidrule(r){6-7} \cmidrule(r){8-8} \cmidrule(r){9-9} 
&&& x-sub & x-view & x-sub & x-setup & x-sub & x-sub \\
\cmidrule{1-9} 
LongT GAN (AAAI'18) &  encoder-decoder & GRU & 52.1 & 56.4 & - & - & 67.7 & 26.5\\
MS$^2$L (ACM MM'20) &  hybrid & GRU & 52.6 & - & - & - & 64.9 & 27.6 \\
PCRP (TMM'21) & hybrid & GRU & \blue{54.9} & 63.4 & 43.0 & 44.6 & - & -\\
AS-CAL (Information Sciences'21) &  contrastive-learning & LSTM & 58.5 & 64.8 & 48.6 & 49.2 & - & -\\ 
EnGAN-PoseRNN (WACV'19) &  encoder-decoder & LSTM & 68.6 & 77.8 & - & - & - & - \\
H-Transformer (ICME'21) & encoder-decoder & Transformer & 69.3 & 72.8 & - & - & - & - \\
SeBiReNet (ECCV'20) &  encoder-decoder & GRU & - & 79.7 & - & - & - & - \\
CrosSCLR (CVPR'21) &  contrastive-learning & GCN & 72.9 & 79.9 & - & - & 84.9* & 21.2* \\
AimCLR (AAAI'22) &  contrastive-learning & GCN & 74.3 & 79.7 & 63.4 & 63.4 & 87.8* & 38.5*\\
Colorization (ICCV'21) &  encoder-decoder & \blue{GCN} & 75.2 & 83.1 & - & - & - & - \\
GL-Transformer (ECCV'22) &  encoder-decoder & Transformer & 76.3 & 83.8 & 66.0 & 68.7 & - & - \\
ISC (ACM MM'21) &  contrastive-learning & GRU\&CNN\&GCN & 76.3 & 85.2 & 67.1 & 67.9 & 80.9 & 36.0  \\
\cmidrule{1-9} 
HiCo-GRU (ours) &  contrastive-learning & GRU & \textbf{80.6} & \textbf{88.6} & \textbf{72.5} & \textbf{73.8} & \textbf{88.9} & \textbf{52.2}\\ 
HiCo-LSTM (ours) &  contrastive-learning & LSTM & \textbf{81.4} & \textbf{88.8} & \textbf{73.7} & \textbf{74.5} & \textbf{89.4} & \textbf{54.7}\\ 
HiCo-Transformer (ours) &  contrastive-learning & Transformer & \textbf{81.1} & \textbf{88.6} & \textbf{72.8} & \textbf{74.1} & \textbf{89.3} & \blue{\textbf{49.4}}\\ 
\bottomrule
\end{tabular}
} 
\end{table*}

\subsection{Comparison to the State-of-the-art}\label{ssec:sota}
In this section, we compare our HiCo to the recently proposed unsupervised state-of-the-art methods, including:\\
(1) Encoder-decoder methods: LongT GAN~\cite{zheng2018unsupervised}, EnGAN-PoseRNN \cite{kundu2019unsupervised}, P\&C \cite{su2020predict}, SeBiReNet \cite{nie2020unsupervised}, H-Transformer \cite{cheng2021hierarchical}, Colorization \cite{yang2021skeleton}, GL-Transformer \cite{kim2022global}.
(2) Contrastive-learning methods: AS-CAL \cite{rao2021augmented}, CrosSCLR \cite{li20213d}, ISC \cite{thoker2021skeleton}, AimCLR \cite{guo2022contrastive}.
(3) Hybrid methods: MS$^2$L \cite{lin2020ms2l}, PCRP \cite{xu2021prototypical}.

The evaluations are conducted in the context of two downstream tasks, \ie skeleton-based action recognition, and skeleton-based action retrieval, which are commonly adopted in previous works \cite{lin2020ms2l,li20213d,thoker2021skeleton}.
For both two downstream tasks, each model should be first trained in an unsupervised manner without using any labeled data, and then used for further evaluation.

\subsubsection{Skeleton-based Action Recognition.}
In this task, an extra linear classifier (a fully connected layer) is further added on the skeleton sequence representation obtained by the corresponding pre-training model, which is trained on the target dataset. Following the previous works \cite{li20213d, lin2020ms2l, guo2022contrastive}, the pre-training model is frozen, and only the linear classifier will be trained.

Table \ref{tab:sota-unsupervised} summarizes the performance on NTU-60, NTU-120, PKU-MMD I and II.
Our proposed HiCo models consistently outperform previous methods of all types with a large margin.
Among the four datasets, all models on PKU-MMD II perform the worst, which is due to that this dataset is more challenging with more noise caused by the view variation~\cite{guo2022contrastive}. Even on this challenging dataset, our best model surpasses the state-of-the-art, \ie AimCLR, by 16\%. 
The results verify the effectiveness of our proposed framework for unsupervised SARL.
In more detail, for our three HiCo variants with different S2S encoders, their performance is comparable, showing a good generality of our framework for S2S encoders.
Additionally, for the compared contrastive-learning methods, they represent a skeleton sequence into instance-level features and perform contrast holistically.
Differently, we represent it into multiple-level features and perform contrast hierarchically, which enables the network to perceive more fine-grained supervision. Hence, it gains large improvements compared to the existing contrastive-learning methods.

\begin{figure}[tb!]
\centering\includegraphics[width=0.92\columnwidth]{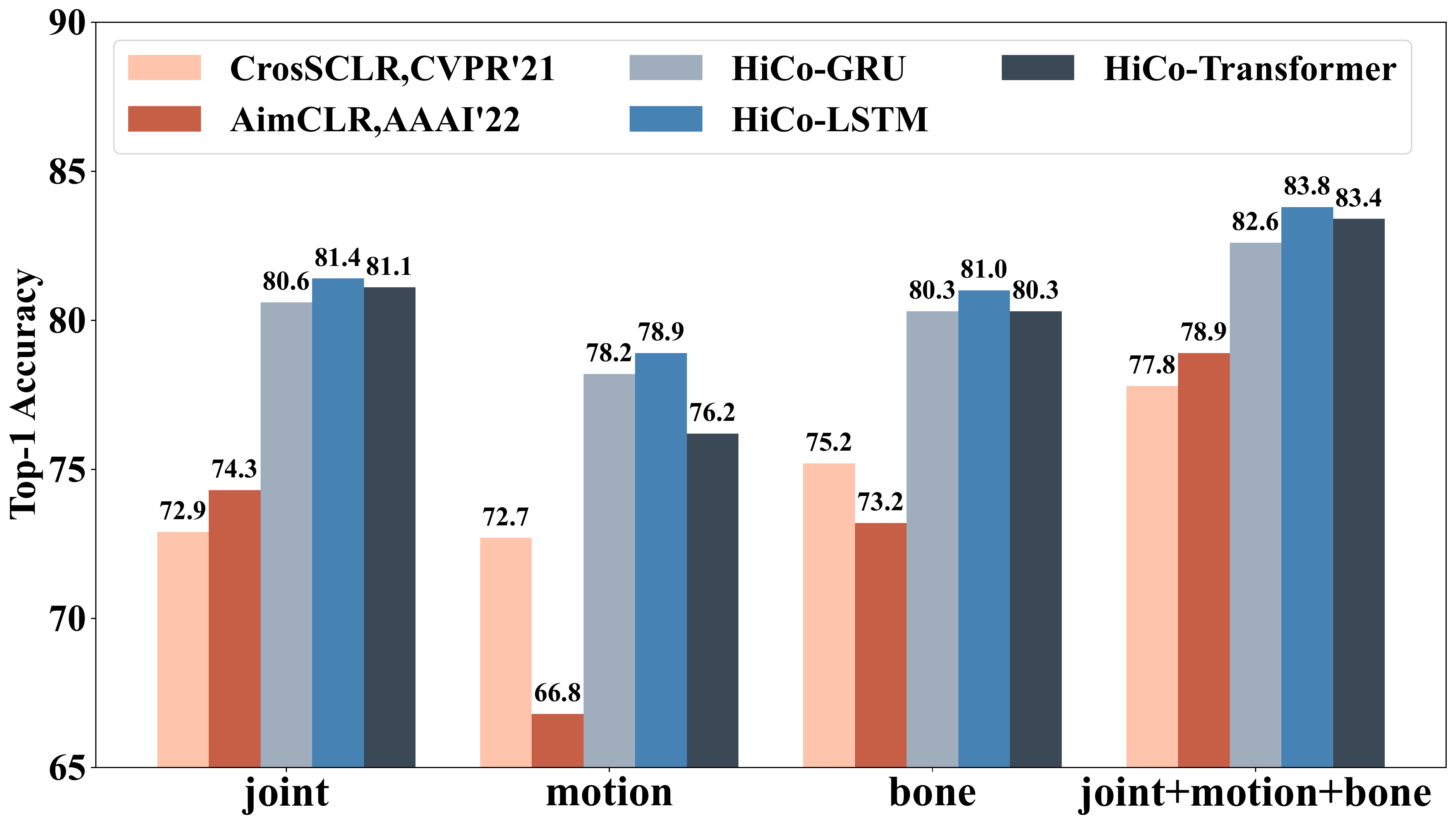}
\caption{Performance comparison using different views of skeleton sequences in terms of x-sub on NTU-60.} \label{fig:ntu60sub}
\end{figure}

Until now we only report our performance using the joint view of skeletons as input.
For a more comprehensive comparison, we also try other views of skeleton sequences and three-streams fusion version of joints, motions, and bones as previous works~\cite{li20213d,guo2022contrastive}. For fusion version, we follow the setting of \cite{li20213d}.
The results are summarized in Figure \ref{fig:ntu60sub}.
Our models consistently outperform the counterparts with a clear margin, which further verifies the effectiveness of our framework. Besides, jointly using three views as input achieves a performance gain.

\begin{table} [tb!]
\renewcommand{\arraystretch}{1.2}
\caption{Comparisons to the state-of-the-art methods for skeleton-based action retrieval on NTU-60 and NTU-120.
}
\label{tab:sota-knn}
\centering 
\scalebox{0.8}{
\begin{tabular}{@{}l*{5}c @{}}
\toprule
\multirow{2}{*}{\textbf{Method}}   & 
\multicolumn{2}{c}{\textbf{NTU-60}} &
\multicolumn{2}{c}{\textbf{NTU-120}} \\
\cmidrule(r){2-3} \cmidrule(r){4-5} 
& x-sub & x-view & x-sub & x-setup & \\
\cmidrule{1-6}
LongT GAN (AAAI'18) & 39.1 & 48.1 & 31.5 & 35.5 & \\
P\&C (CVPR'20) & 50.7 & 76.3 & 39.5 & 41.8 & \\
AimCLR (AAAI'22) & 62.0 & - & - & - & \\
ISC (ACMMM'21) & 62.5 & 82.6 & 50.6 & 52.3 & \\
\hline
HiCo-GRU (ours) & \textbf{67.9} & \textbf{84.4} & \textbf{55.9} & \textbf{58.7}  \\
HiCo-LSTM (ours) & \textbf{66.9} & \textbf{84.3} & \textbf{56.3} & \textbf{59.1}  \\
HiCo-Transformer (ours) & \textbf{68.3} & \textbf{84.8} & \textbf{56.6} & \textbf{59.1}   \\
\bottomrule
\end{tabular}
}
\end{table}

\subsubsection{Skeleton-based Action Retrieval.}
In this experiment, we follow the setting in \cite{ thoker2021skeleton, guo2022contrastive}. Specifically, given an action query, the most similar action is retrieved from training samples using cosine similarity. 
Table \ref{tab:sota-knn} shows the results on NTU-60 and NTU-120. Again, our HiCo models with various encoders consistently perform better than the existing methods. 
The result further demonstrates that the action representation obtained by our HiCo is more discriminative.

\begin{figure}[tb!] 
\centering
\subfigure[Temporal branch]{
\begin{minipage}[t]{0.48\linewidth}
\centering
\includegraphics[width=1.5in]{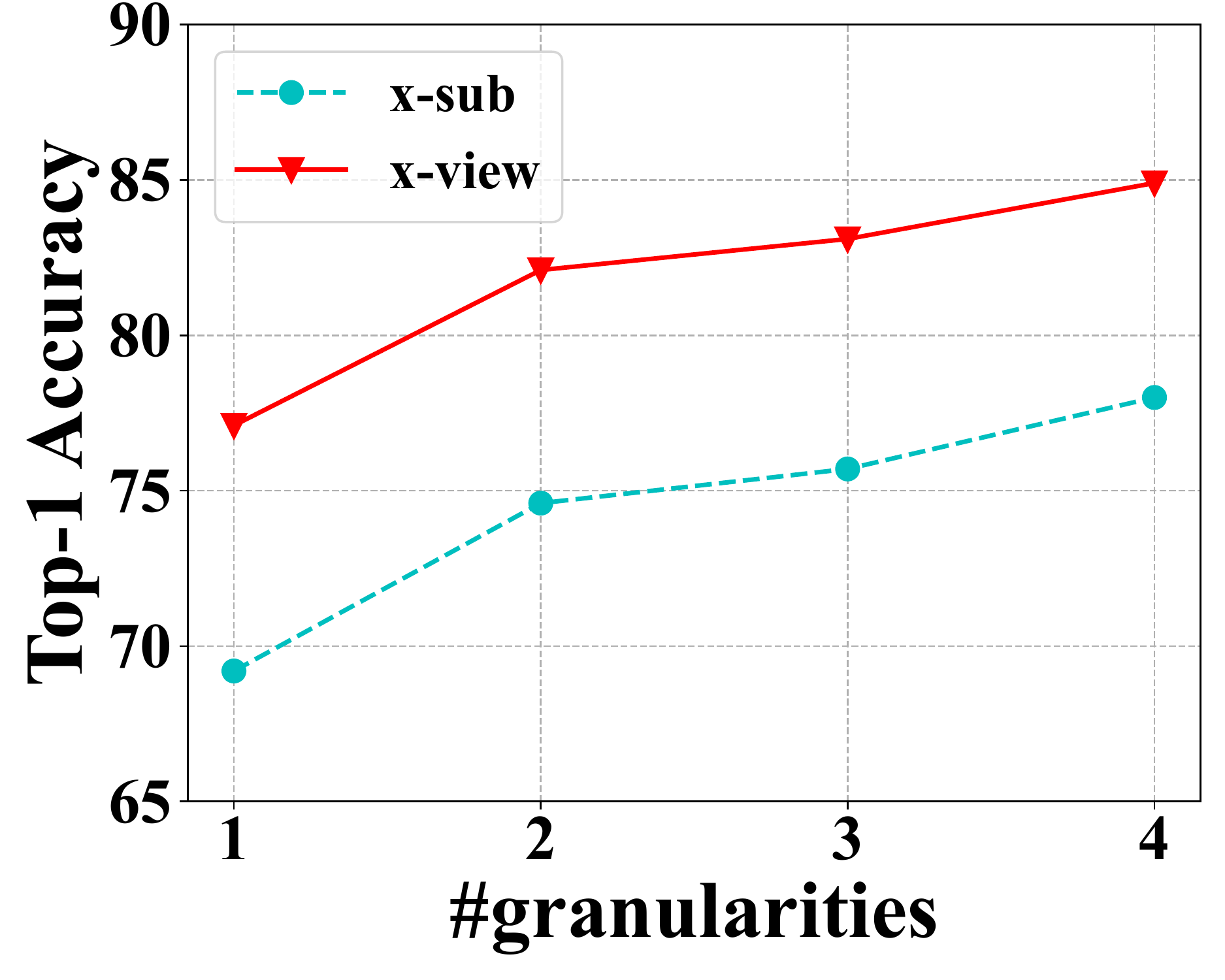}
\end{minipage}%
}
\subfigure[Spatial branch]{
\begin{minipage}[t]{0.48\linewidth}
\centering
\includegraphics[width=1.5in]{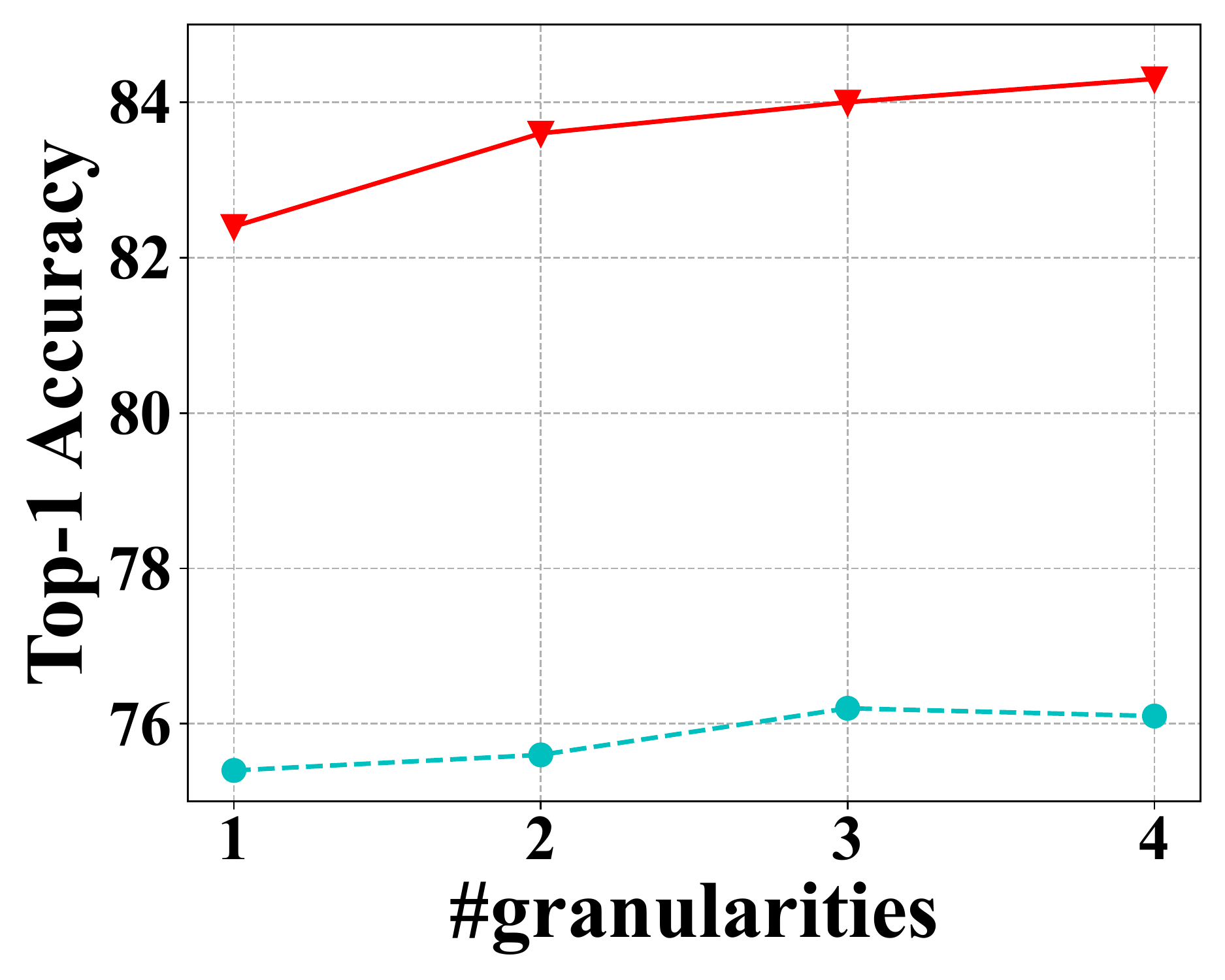}
\end{minipage}%
}
\centering
\caption{The effectiveness of representation from multiple temporal and spatial granularities.
The performance is gradually increased with more granularities being modeled.
}\label{fig:granularity}
\end{figure}

\subsection{Ablation \blue{Studies}}\label{ssec:ablation}

To verify the effectiveness of each component in our proposed framework, we conduct ablation studies on NTU-60. All the experiments are conducted in the context of the skeleton-based action recognition downstream task, and Transformer is used as the default encoder of our model.

\textbf{The effectiveness of representation on multiple granularities.}
Figure \ref{fig:granularity} illustrates the performance of our models using the different numbers of granularities.
For both branches, the performance is gradually increased with more granularities being added. The results show the effectiveness of modeling multiple temporal and spatial granularities.

\begin{table} [tb!]
\renewcommand{\arraystretch}{1.5}
\caption{The effectiveness of two-branch structure that jointly models temporal and spatial dependencies.
}
\label{tab:ablation-branches}
\centering 
\scalebox{0.9}{
\begin{tabular}{*{2}c|*{2}c}
\toprule
\textbf{Temporal branch} & \textbf{Spatial branch} & \textbf{x-sub} & \textbf{x-view} \\
\hline
\checkmark & - & 78.0 & 84.9  \\
-  & \checkmark & 76.1 & 84.3  \\
\checkmark & \checkmark & \textbf{79.8} & \textbf{87.0}  \\
\bottomrule
\end{tabular}
}
\end{table}
\begin{table} [tb!]
\renewcommand{\arraystretch}{1.5}
\caption{The effectiveness of hierarchical contrast on multi-level features.
\blue{All variants are based on the same representation of multiple granularities, but utilizing different losses.}
}
\label{tab:ablation-losses}
\centering 
\scalebox{0.8}{
\begin{tabular}{*{3}c|*{2}c}
\toprule
\textbf{Instance level} & \textbf{Domain level} & \textbf{Clip\&Part level} & \textbf{x-sub} & \textbf{x-view} \\
\hline
\checkmark & - & - & 79.8 & 87.0  \\
\checkmark & \checkmark & - & 80.7 & 88.4  \\
\checkmark & \checkmark & \checkmark & \textbf{81.1} & \textbf{88.6}  \\
\bottomrule
\end{tabular}
}
\end{table}

\textbf{The effectiveness of two-branch structure.}
Table \ref{tab:ablation-branches} shows the results of our model using one branch or two branches. The two-branch consistently performs better than the one-branch counterparts.
It not only demonstrates the effectiveness of the two-branch structure, but also shows the complementary of the temporal and the spatial branches.

\textbf{The effectiveness of hierarchical contrast.}
Table \ref{tab:ablation-losses} summarizes the performance of conducting contrast over features of different levels, and all models use the hierarchical encoder network. It is worth noting that the model only with the instance-level contrast utilizes the same contrastive learning way of AS-CAL \cite{rao2021augmented}, CrosSCLR \cite{li20213d}. Progressively adding domain-level contrast, and clip\&part-level contrasts gain clear improvement, which demonstrates the effectiveness of hierarchical contrast.

\subsection{Transfer Learning for Action Recognition}\label{ssec:transfer}
To explore the transferability of our learned representation, we evaluate whether it learned from a source dataset generalizes well to another target dataset.
Specifically, a model is first pre-trained on a source dataset in an unsupervised manner, then finetuned on a target dataset.
For a cross-paper comparison, following \cite{thoker2021skeleton,lin2020ms2l}, we choose NTU-60 and PKU-MMD I as the source datasets, and PKU-MMD II as the target dataset. All the results are evaluated under the x-sub protocol.
As shown in Table \ref{tab:sota-transfer}, our proposed HiCo outperforms its competitors with a large margin, which shows good transferability of our proposed action representation learning for skeleton-based action recognition. 

\begin{table} [tb!]
\renewcommand{\arraystretch}{1.2}
\caption{Comparisons on transfer learning.
}
\label{tab:sota-transfer}
\centering 
\scalebox{0.8}{
\begin{tabular}{@{}l*{3}c @{}}
\toprule
\multirow{2}{*}{\textbf{Method}}   & 
\multicolumn{2}{c}{\textbf{Transfer to PKU-MMD II}} &\\
\cmidrule{2-3} 
& \multicolumn{1}{c}{\textbf{PKU-MMD I}} &
\multicolumn{1}{c}{\textbf{NTU-60}} &\\
\cmidrule{2-3} 
\cmidrule{1-4}
LongT GAN \cite{zheng2018unsupervised} & 43.6 & 44.8 &  \\
M$^2$L \cite{lin2020ms2l} & 44.1 & 45.8 &  \\
ISC \cite{thoker2021skeleton} & 45.1 & 45.9 &  \\
\cmidrule{1-4}
HiCo-Transformer (ours)  & \textbf{53.4} & \textbf{56.3}   \\
\bottomrule
\end{tabular}
 }
\end{table}
\begin{table} [tb!]
\renewcommand{\arraystretch}{1.2}
\caption{Comparisons to the state-of-the-art methods with semi-supervised learning on NTU-60 dataset. 
}
\label{tab:sota-semisupervised}
%
\centering 
\scalebox{0.8}{
\begin{tabular}{@{}l*{5}c @{}}
\toprule
\multirow{2}{*}{\textbf{Method}}   &
\multicolumn{2}{c}{\textbf{x-sub}} & \multicolumn{2}{c}{\textbf{x-view}} \\

\cmidrule(r){2-3} \cmidrule(r){4-5} 
& 1\%  & 10\%  & 1\%  & 10\%  \\
\cmidrule{1-5}
LongGAN \cite{zheng2018unsupervised} & 35.2 & 62.0 & - & -    \\
M$^2$L \cite{lin2020ms2l} & 33.1 & 65.2 & - & -    \\
\blue{ASSL \cite{si2020adversarial}} & \blue{-} & \blue{64.3} & \blue{-} & \blue{69.8}   \\
ISC \cite{thoker2021skeleton} & 35.7 & 65.9 & 38.1 & 72.5  \\
MCC \cite{su2021modeling} & - & 60.8 & - & 65.8 \\
Colorization \cite{yang2021skeleton} & 48.3 & 71.7 & 52.5 & \textbf{78.9}  \\
CrosSCLR \cite{li20213d} & - & 67.6 & - & 73.5 \\
Hi-TRS \cite{chen2022hierarchically} & - & 70.7 & - & 74.8 \\
\blue{GL-Transformer} \cite{kim2022global} & - & 68.6 & - & 74.9 \\
\hline
HiCo-Transformer (Ours) & \textbf{54.4} & \textbf{73.0} &  \textbf{54.8} &  78.3   \\
\bottomrule
\end{tabular}
 }
\end{table}

\subsection{The Potential for Semi-Supervised Learning}\label{ssec:semi}

For semi-supervised learning, we follow the setting of \cite{lin2020ms2l,chen2022hierarchically},
and report the results of using 1\% and 10\% randomly sampled training data with labels for finetuning.
Table \ref{tab:sota-semisupervised} summarizes the results on NTU-60.
Our HiCo with Transformer as encoders performs the best on the setting of 1\% data, and is slightly worse than Colorization~\cite{yang2021skeleton} on the setting of 10\% data. 
\blue{We attribute the better performance of Colorization to that it has larger trainable parameters (10.9M) than ours (4.1M).
}
The results allow us to conclude that our HiCo is more effective when fewer labeled training data are available. We attribute it to our hierarchical contrast on multiple-level representation, which enables the network to acquire more supervision thus be more effective in the scenario of fewer labeled data.

\section{Conclusion} \label{sec:conc}
This paper proposes a new contrastive-learning framework HiCo for unsupervised SARL.
By jointly exploiting multiple-level representation and hierarchical contrast, HiCo is capable of learning more discriminative representation under unsupervised scenarios and has good transferability.
Extensive experiments show the effectiveness of HiCo and its new state-of-the-art performance.
Given HiCo is simple and effective, we believe it can also be used as a new strong baseline for unsupervised SARL.

\section*{Acknowledgments}
This work was supported by the NSFC (61902347, 62002323, 61976188), the Public Welfare Technology Research Project of Zhejiang Province (LGF21F020010), the Open Projects Program of the National Laboratory of Pattern Recognition, the Fundamental Research Funds for the Provincial Universities of Zhejiang.

\bibliography{aaai23}

\begin{thebibliography}{40}
\providecommand{\natexlab}[1]{#1}

\bibitem[{Chen et~al.(2020)Chen, Kornblith, Norouzi, and
  Hinton}]{chen2020simple}
Chen, T.; Kornblith, S.; Norouzi, M.; and Hinton, G. 2020.
\newblock A simple framework for contrastive learning of visual
  representations.
\newblock In \emph{International Conference on Machine Learning}, 1597--1607.

\bibitem[{Chen et~al.(2022)Chen, Zhao, Yuan, Tian, Xia, Geng, Han, and
  Metaxas}]{chen2022hierarchically}
Chen, Y.; Zhao, L.; Yuan, J.; Tian, Y.; Xia, Z.; Geng, S.; Han, L.; and
  Metaxas, D.~N. 2022.
\newblock Hierarchically self-supervised transformer for human skeleton
  representation learning.
\newblock In \emph{European Conference on Computer Vision}, 185--202.

\bibitem[{Cheng et~al.(2020)Cheng, Zhang, He, Chen, Cheng, and
  Lu}]{cheng2020skeleton}
Cheng, K.; Zhang, Y.; He, X.; Chen, W.; Cheng, J.; and Lu, H. 2020.
\newblock Skeleton-based action recognition with shift graph convolutional
  network.
\newblock In \emph{Proceedings of the IEEE/CVF Conference on Computer Vision
  and Pattern Recognition}, 183--192.

\bibitem[{Cheng et~al.(2021)Cheng, Chen, Chen, Wei, Zhang, and
  Lin}]{cheng2021hierarchical}
Cheng, Y.-B.; Chen, X.; Chen, J.; Wei, P.; Zhang, D.; and Lin, L. 2021.
\newblock Hierarchical transformer: Unsupervised representation learning for
  skeleton-based human action recognition.
\newblock In \emph{2021 IEEE International Conference on Multimedia and Expo
  (ICME)}, 1--6.

\bibitem[{Davies and Bouldin(1979)}]{davies1979cluster}
Davies, D.~L.; and Bouldin, D.~W. 1979.
\newblock A cluster separation measure.
\newblock \emph{IEEE Transactions on Pattern Analysis and Machine
  Intelligence}, (2): 224--227.

\bibitem[{Deng et~al.(2019)Deng, Yang, Liu, Li, Liu, and
  Tao}]{deng2019unsupervised}
Deng, C.; Yang, E.; Liu, T.; Li, J.; Liu, W.; and Tao, D. 2019.
\newblock Unsupervised semantic-preserving adversarial hashing for image
  search.
\newblock \emph{IEEE Transactions on Image Processing}, 28(8): 4032--4044.

\bibitem[{Dong et~al.(2022)Dong, Li, Xu, Yang, Yang, Wang, and
  Wang}]{dong2021dual}
Dong, J.; Li, X.; Xu, C.; Yang, X.; Yang, G.; Wang, X.; and Wang, M. 2022.
\newblock Dual encoding for video retrieval by text.
\newblock \emph{IEEE Transactions on Pattern Analysis and Machine
  Intelligence}, 44(8): 4065--4080.

\bibitem[{Du, Wang, and Wang(2015)}]{du2015hierarchical}
Du, Y.; Wang, W.; and Wang, L. 2015.
\newblock Hierarchical recurrent neural network for skeleton based action
  recognition.
\newblock In \emph{Proceedings of the IEEE Conference on Computer Vision and
  Pattern Recognition}, 1110--1118.

\bibitem[{Gao et~al.(2020)Gao, Yang, Zhang, and Xu}]{gao2020unsupervised}
Gao, J.; Yang, X.; Zhang, Y.; and Xu, C. 2020.
\newblock Unsupervised video summarization via relation-aware assignment
  learning.
\newblock \emph{IEEE Transactions on Multimedia}, 23: 3203--3214.

\bibitem[{Gao, Zhang, and Xu(2019)}]{gao2019know}
Gao, J.; Zhang, T.; and Xu, C. 2019.
\newblock I know the relationships: Zero-shot action recognition via two-stream
  graph convolutional networks and knowledge graphs.
\newblock In \emph{Proceedings of the AAAI conference on artificial
  intelligence}, volume~33, 8303--8311.

\bibitem[{Guo et~al.(2022)Guo, Liu, Chen, Liu, Wang, and
  Ding}]{guo2022contrastive}
Guo, T.; Liu, H.; Chen, Z.; Liu, M.; Wang, T.; and Ding, R. 2022.
\newblock Contrastive learning from extremely augmented skeleton sequences for
  self-supervised action recognition.
\newblock In \emph{Proceedings of the AAAI Conference on Artificial
  Intelligence}, volume~36, 762--770.

\bibitem[{He et~al.(2020)He, Fan, Wu, Xie, and Girshick}]{he2020momentum}
He, K.; Fan, H.; Wu, Y.; Xie, S.; and Girshick, R. 2020.
\newblock Momentum contrast for unsupervised visual representation learning.
\newblock In \emph{Proceedings of the IEEE/CVF Conference on Computer Vision
  and Pattern Recognition}, 9729--9738.

\bibitem[{Kim et~al.(2022)Kim, Chang, Kim, and Choi}]{kim2022global}
Kim, B.; Chang, H.~J.; Kim, J.; and Choi, J.~Y. 2022.
\newblock Global-local motion transformer for unsupervised skeleton-based
  action learning.
\newblock In \emph{European Conference on Computer Vision}, 209--225.

\bibitem[{Kundu et~al.(2019)Kundu, Gor, Uppala, and
  Radhakrishnan}]{kundu2019unsupervised}
Kundu, J.~N.; Gor, M.; Uppala, P.~K.; and Radhakrishnan, V.~B. 2019.
\newblock Unsupervised feature learning of human actions as trajectories in
  pose embedding manifold.
\newblock In \emph{2019 IEEE Winter Conference on Applications of Computer
  Vision (WACV)}, 1459--1467.

\bibitem[{Li et~al.(2021)Li, Wang, Ni, Wang, Yang, and Zhang}]{li20213d}
Li, L.; Wang, M.; Ni, B.; Wang, H.; Yang, J.; and Zhang, W. 2021.
\newblock 3D human action representation learning via cross-view consistency
  pursuit.
\newblock In \emph{Proceedings of the IEEE/CVF Conference on Computer Vision
  and Pattern Recognition}, 4741--4750.

\bibitem[{Li et~al.(2019)Li, Xu, Yang, Chen, and Dong}]{li2019w2vv++}
Li, X.; Xu, C.; Yang, G.; Chen, Z.; and Dong, J. 2019.
\newblock {W2VV++} fully deep learning for ad-hoc video search.
\newblock In \emph{Proceedings of the 27th ACM International Conference on
  Multimedia}, 1786--1794.

\bibitem[{Lin et~al.(2020)Lin, Song, Yang, and Liu}]{lin2020ms2l}
Lin, L.; Song, S.; Yang, W.; and Liu, J. 2020.
\newblock Ms2l: Multi-task self-supervised learning for skeleton based action
  recognition.
\newblock In \emph{Proceedings of the 28th ACM International Conference on
  Multimedia}, 2490--2498.

\bibitem[{Liu et~al.(2022)Liu, Qu, Wang, Di, Zou, Cheng, Xu, and
  Zhou}]{liu2022unsupervised}
Liu, D.; Qu, X.; Wang, Y.; Di, X.; Zou, K.; Cheng, Y.; Xu, Z.; and Zhou, P.
  2022.
\newblock Unsupervised temporal video grounding with deep semantic clustering.
\newblock 1683--1691.

\bibitem[{Liu et~al.(2019)Liu, Shahroudy, Perez, Wang, Duan, and
  Kot}]{liu2019ntu}
Liu, J.; Shahroudy, A.; Perez, M.; Wang, G.; Duan, L.-Y.; and Kot, A.~C. 2019.
\newblock NTU RGB+D 120: A large-scale benchmark for 3D human activity
  understanding.
\newblock \emph{IEEE Transactions on Pattern Analysis and Machine
  Intelligence}, 42(10): 2684--2701.

\bibitem[{Liu et~al.(2020)Liu, Song, Liu, Li, and Hu}]{liu2020benchmark}
Liu, J.; Song, S.; Liu, C.; Li, Y.; and Hu, Y. 2020.
\newblock A benchmark dataset and comparison study for multi-modal human action
  analytics.
\newblock \emph{ACM Transactions on Multimedia Computing, Communications, and
  Applications (TOMM)}, 16(2): 1--24.

\bibitem[{Ni, Wang, and Moulin(2011)}]{ni2011rgbd}
Ni, B.; Wang, G.; and Moulin, P. 2011.
\newblock RGBD-HuDaAct: A color-depth video database for human daily activity
  recognition.
\newblock In \emph{IEEE International Conference on Computer Vision Workshops
  (ICCV Workshops)}, 1147--1153.

\bibitem[{Nie, Liu, and Liu(2020)}]{nie2020unsupervised}
Nie, Q.; Liu, Z.; and Liu, Y. 2020.
\newblock Unsupervised 3D human pose representation with viewpoint and pose
  disentanglement.
\newblock In \emph{European Conference on Computer Vision}, 102--118.

\bibitem[{Rao et~al.(2021)Rao, Xu, Hu, Cheng, and Hu}]{rao2021augmented}
Rao, H.; Xu, S.; Hu, X.; Cheng, J.; and Hu, B. 2021.
\newblock Augmented skeleton based contrastive action learning with momentum
  lstm for unsupervised action recognition.
\newblock \emph{Information Sciences}, 569: 90--109.

\bibitem[{Shahroudy et~al.(2016)Shahroudy, Liu, Ng, and
  Wang}]{shahroudy2016ntu}
Shahroudy, A.; Liu, J.; Ng, T.-T.; and Wang, G. 2016.
\newblock NTU RGB+D: A large scale dataset for 3D human activity analysis.
\newblock In \emph{Proceedings of the IEEE Conference on Computer Vision and
  Pattern Recognition}, 1010--1019.

\bibitem[{Si et~al.(2020)Si, Nie, Wang, Wang, Tan, and
  Feng}]{si2020adversarial}
Si, C.; Nie, X.; Wang, W.; Wang, L.; Tan, T.; and Feng, J. 2020.
\newblock Adversarial self-supervised learning for semi-supervised 3D action
  recognition.
\newblock In \emph{European Conference on Computer Vision}, 35--51.

\bibitem[{Su, Liu, and Shlizerman(2020)}]{su2020predict}
Su, K.; Liu, X.; and Shlizerman, E. 2020.
\newblock Predict \& cluster: Unsupervised skeleton based action recognition.
\newblock In \emph{Proceedings of the IEEE/CVF Conference on Computer Vision
  and Pattern Recognition}, 9631--9640.

\bibitem[{Su et~al.(2021)Su, Lin, Sun, Hao, and Wu}]{su2021modeling}
Su, Y.; Lin, G.; Sun, R.; Hao, Y.; and Wu, Q. 2021.
\newblock Modeling the uncertainty for self-supervised 3D skeleton action
  representation learning.
\newblock In \emph{Proceedings of the 29th ACM International Conference on
  Multimedia}, 769--778.

\bibitem[{Su, Lin, and Wu(2021)}]{su2021self}
Su, Y.; Lin, G.; and Wu, Q. 2021.
\newblock Self-supervised 3D skeleton action representation learning with
  motion consistency and continuity.
\newblock In \emph{Proceedings of the IEEE/CVF International Conference on
  Computer Vision}, 13328--13338.

\bibitem[{Thoker, Doughty, and Snoek(2021)}]{thoker2021skeleton}
Thoker, F.~M.; Doughty, H.; and Snoek, C.~G. 2021.
\newblock Skeleton-contrastive 3D action representation learning.
\newblock In \emph{Proceedings of the 29th ACM International Conference on
  Multimedia}, 1655--1663.

\bibitem[{Van~den Oord et~al.(2018)Van~den Oord, Li, Vinyals
  et~al.}]{van2018representation}
Van~den Oord, A.; Li, Y.; Vinyals, O.; et~al. 2018.
\newblock Representation learning with contrastive predictive coding.
\newblock \emph{arXiv preprint arXiv:1807.03748}.

\bibitem[{Van~der Maaten and Hinton(2008)}]{van2008visualizing}
Van~der Maaten, L.; and Hinton, G. 2008.
\newblock Visualizing data using t-SNE.
\newblock \emph{Journal of Machine Learning Research}, 9(11).

\bibitem[{Vemulapalli and Chellapa(2016)}]{vemulapalli2016rolling}
Vemulapalli, R.; and Chellapa, R. 2016.
\newblock Rolling rotations for recognizing human actions from 3D skeletal
  data.
\newblock In \emph{Proceedings of the IEEE Conference on Computer Vision and
  Pattern Recognition}, 4471--4479.

\bibitem[{Wang et~al.(2012)Wang, Liu, Wu, and Yuan}]{wang2012mining}
Wang, J.; Liu, Z.; Wu, Y.; and Yuan, J. 2012.
\newblock Mining actionlet ensemble for action recognition with depth cameras.
\newblock In \emph{2012 IEEE Conference on Computer Vision and Pattern
  Recognition}, 1290--1297.

\bibitem[{Wang, Ni, and Yang(2020)}]{wang2020learning}
Wang, M.; Ni, B.; and Yang, X. 2020.
\newblock Learning multi-view interactional skeleton graph for action
  recognition.
\newblock \emph{IEEE Transactions on Pattern Analysis and Machine
  Intelligence}.

\bibitem[{Xu et~al.(2021)Xu, Rao, Hu, Cheng, and Hu}]{xu2021prototypical}
Xu, S.; Rao, H.; Hu, X.; Cheng, J.; and Hu, B. 2021.
\newblock Prototypical contrast and reverse prediction: Unsupervised skeleton
  based action recognition.
\newblock \emph{IEEE Transactions on Multimedia}, 1--1.

\bibitem[{Yang et~al.(2018)Yang, Deng, Liu, Liu, and Tao}]{yang2018semantic}
Yang, E.; Deng, C.; Liu, T.; Liu, W.; and Tao, D. 2018.
\newblock Semantic structure-based unsupervised deep hashing.
\newblock In \emph{Proceedings of the 27th International Joint Conference on
  Artificial Intelligence}, 1064--1070.

\bibitem[{Yang et~al.(2021)Yang, Liu, Lu, Er, and Kot}]{yang2021skeleton}
Yang, S.; Liu, J.; Lu, S.; Er, M.~H.; and Kot, A.~C. 2021.
\newblock Skeleton cloud colorization for unsupervised 3D action representation
  learning.
\newblock In \emph{Proceedings of the IEEE/CVF International Conference on
  Computer Vision}, 13423--13433.

\bibitem[{Zhang et~al.(2017)Zhang, Lan, Xing, Zeng, Xue, and
  Zheng}]{zhang2017view}
Zhang, P.; Lan, C.; Xing, J.; Zeng, W.; Xue, J.; and Zheng, N. 2017.
\newblock View adaptive recurrent neural networks for high performance human
  action recognition from skeleton data.
\newblock In \emph{Proceedings of the IEEE International Conference on Computer
  Vision}, 2117--2126.

\bibitem[{Zhang(2012)}]{zhang2012microsoft}
Zhang, Z. 2012.
\newblock Microsoft kinect sensor and its effect.
\newblock \emph{IEEE Multimedia}, 19(2): 4--10.

\bibitem[{Zheng et~al.(2018)Zheng, Wen, Liu, Long, Dai, and
  Gong}]{zheng2018unsupervised}
Zheng, N.; Wen, J.; Liu, R.; Long, L.; Dai, J.; and Gong, Z. 2018.
\newblock Unsupervised representation learning with long-term dynamics for
  skeleton based action recognition.
\newblock In \emph{Proceedings of the AAAI Conference on Artificial
  Intelligence}, volume~32, 2644--2651.

\end{thebibliography}

\clearpage

\appendix
\section*{Supplementary Material}
This supplementary material contains the following contents which are not included in the paper due to space limits: 
\begin{itemize}
    \item Detailed descriptions of the datasets used in our experiment (Section \ref{sec:dataset}).
    \item Additional ablation studies including the influence of different UDM structures (Section \ref{ssec:udm}), and the influence of different fusion ways (Section \ref{ssec:fusion}). 
    \item More experiments including the results using different views as input on NTU-60, NTU-120 adn PKU-MMD I (Section \ref{ssec:cmp}), the complexity comparison (Section \ref{ssec:cost}), and the t-SNE visualization of learned representation (Section \ref{ssec:vr}).
    \item Implementation details including data augmentation, model structure, pretraining details, and training details for downstream tasks (Section \ref{sec:id}).
\end{itemize}

\section{Datasets}\label{sec:dataset}

Following the previous works~\cite{thoker2021skeleton,li20213d,guo2022contrastive}, we utilze four skeleton-based action datasets, \ie NTU-60 \cite{shahroudy2016ntu}, NTU-120 \cite{liu2019ntu}, PKU-MMD I, and PKU-MMD II \cite{liu2020benchmark}.

\textbf{NTU-60} is a skeleton-based action recognition dataset that consists of 56,880 video samples of 60 action categories, which are captured from 40 different human subjects. There are two standard evaluation protocols.
1) Cross-Subject (x-sub): 20 subjects are used as training data, and the rest 20 subjects are used as validation data. 
2) Cross-View (x-view): samples captured by cameras 2 and 3 are used as training data, and samples captured by camera 1 are used as validation data. 

\textbf{NTU-120} is an extended dataset of NTU-60, which has 60 additional action categories and more skeleton samples. It totally has 120 categories and 113,945 samples.
Actions are performed by 106 different human subjects with 32 setups, and each setup denotes the data collection setting of the specific locations of cameras and a specific background. 
There are two standard protocols. 
1) Cross-Subject (x-sub): similar to NTU-60, 53 subjects are used as training data, and the rest 53 subjects are used as validation data. 
2) Cross-Setup (x-setup): setups with even IDs are used as training data, and setups with odd IDs are used as validation data.

\textbf{PKU-MMD I and II} are datasets originally developed for human activities detection, which have been commonly used for skeleton-based action recognition.  PKU-MMD I is an easier version dataset, while PKU-MMD II is the second version of more challenging due to its larger view variation. 
Both datasets contain 51 action categories, and each category is performed by one or two subjects, resulting in 20,000 action samples for PKU-MMD I and 7,000 action samples for PKU-MMD II.
Following the previous works~\cite{lin2020ms2l,thoker2021skeleton,guo2022contrastive}, we also evaluate our method with x-sub protocol on these two datasets.

\section{Additional Ablation Studies}

\subsection{The Influence of UDM Structures}\label{ssec:udm}
\begin{table} [tb!]
\renewcommand{\arraystretch}{1.2}
\caption{Performance of our model with UDM of different structure on NTU-60.
}
\label{tab:udm}

\centering 
\scalebox{1.0}{
\begin{tabular}{@{}l*{3}c @{}}
\toprule
\textbf{Structure} & x-sub & x-view \\ 
\hline
Conv1d + max-pooing & 81.1 & 88.6 \\
Conv1d + mean-pooling & 80.8 & 88.4 \\
Max-pooling & 79.7 & 87.5 \\
Mean-pooling & 78.8 & 87.3 \\
\bottomrule
\end{tabular}
 }
\end{table}
Recall that our Unified Donwsampling Module (UDM) is implemented by a 1D convolution layer (Conv1d) followed by the max-pooling.   
As shown in Table \ref{tab:udm}, we try replacing the max-pooling with the mean-pooling, obtaining a similar performance to the max-pooling counterpart. Additionally, we also remove the Conv1d in the above two models, and find that their performance has consistent drops. 
The results show the importance of the Conv1d for UDM.
With the various structures of UDM, our models consistently outperform the existing best model with x-sub of 76.3 and x-view of 85.2 on NTU-60 \cite{thoker2021skeleton}. It further verifies the effectiveness of our proposed Hierarchical Contrast framework.

\subsection{The Influence of Fusion Ways}\label{ssec:fusion}
\begin{table} [tb!]
\renewcommand{\arraystretch}{1.2}
\caption{Performance of our model using different fusion ways on NUT-60.
}
\label{tab:fusion}

\centering 
\scalebox{1.0}{
\begin{tabular}{@{}l*{3}c @{}}
\toprule
\textbf{Fusion} & x-sub & x-view \\ 
\hline
Element-wise sum & 80.6 & 87.8 \\
Hadamard product & 79.8 & 86.5 \\
Weighted sum & 80.7 & 87.9 \\
Concatenation & 81.1 & 88.6 \\

\bottomrule
\end{tabular}
 }
\end{table}
We investigate different fusion ways of two branches including element-wise sum, Hadamard product, and weighted sum. Note the weights for the weighted sum are learned by a linear layer. The results are shown in Table \ref{tab:fusion}, where the concatenation operation performs the best.

\begin{figure*}[tb!] 
\centering
\subfigure[NTU-60; x-view]{
\begin{minipage}[t]{0.48\linewidth}
\centering
\includegraphics[width=3.2in]{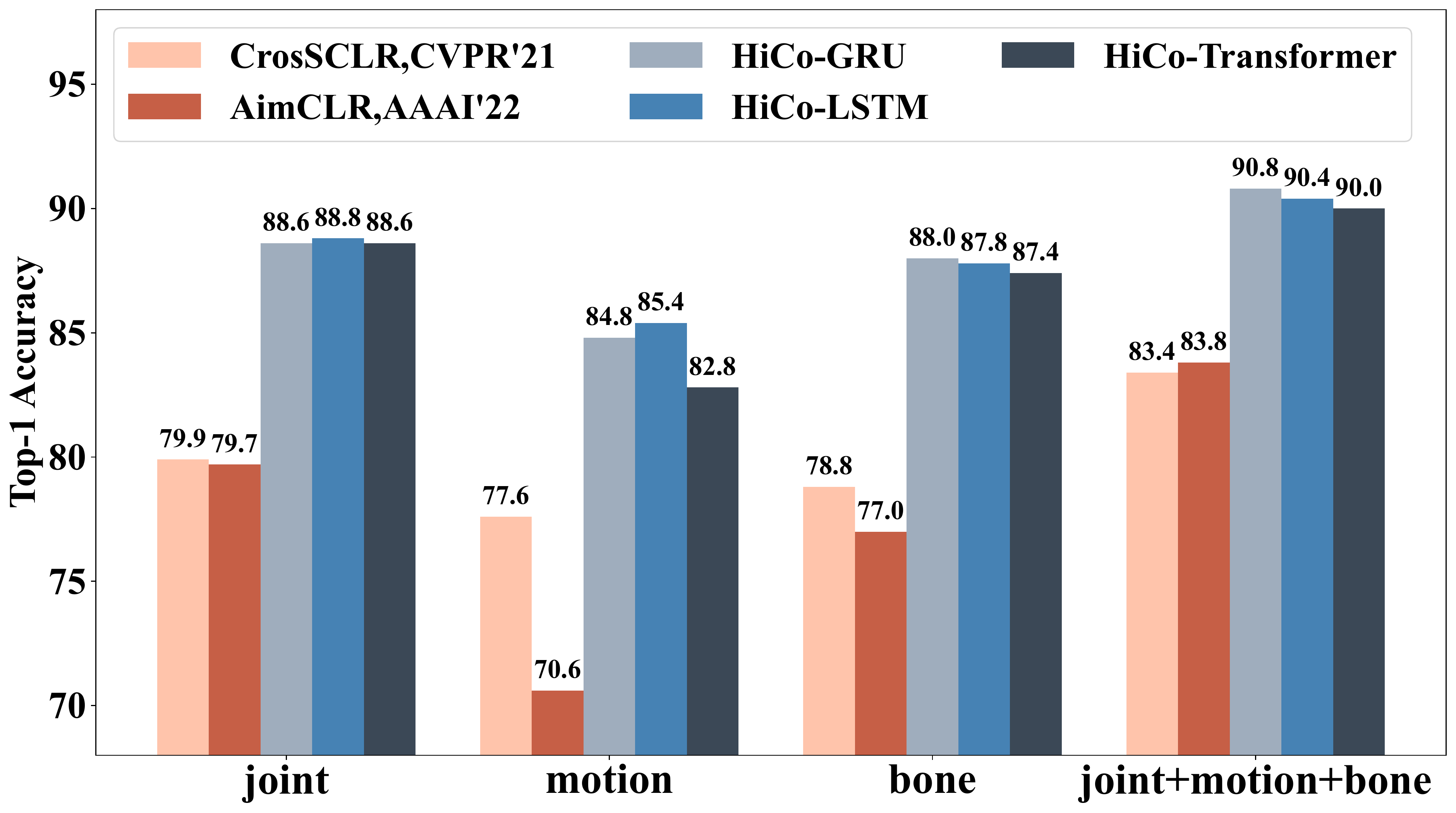}
\end{minipage}%
}
\subfigure[NTU-120; x-sub]{
\begin{minipage}[t]{0.48\linewidth}
\centering
\includegraphics[width=3.2in]{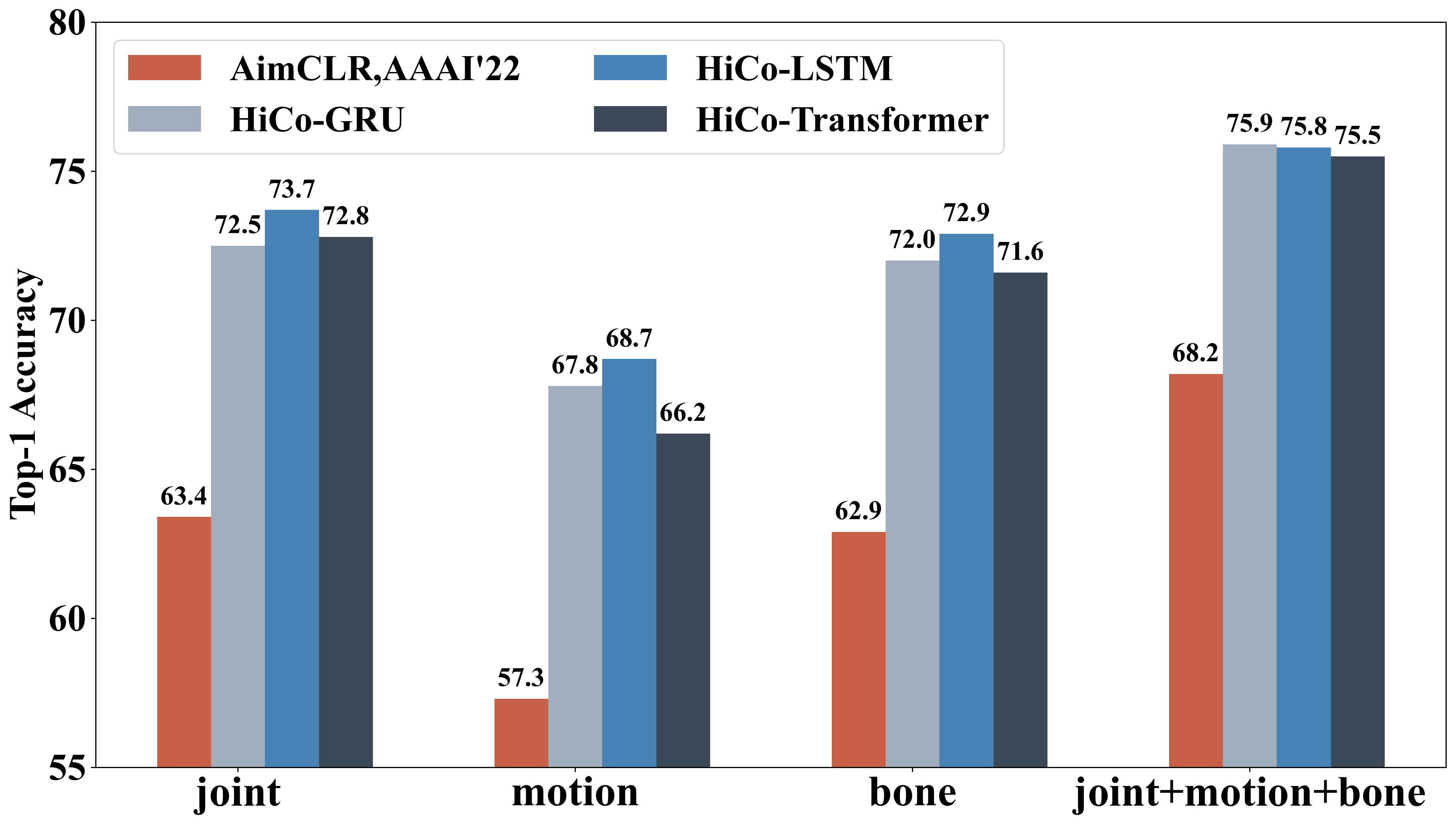}
\end{minipage}%
}
\subfigure[NTU-120; x-set]{
\begin{minipage}[t]{0.48\linewidth}
\centering
\includegraphics[width=3.2in]{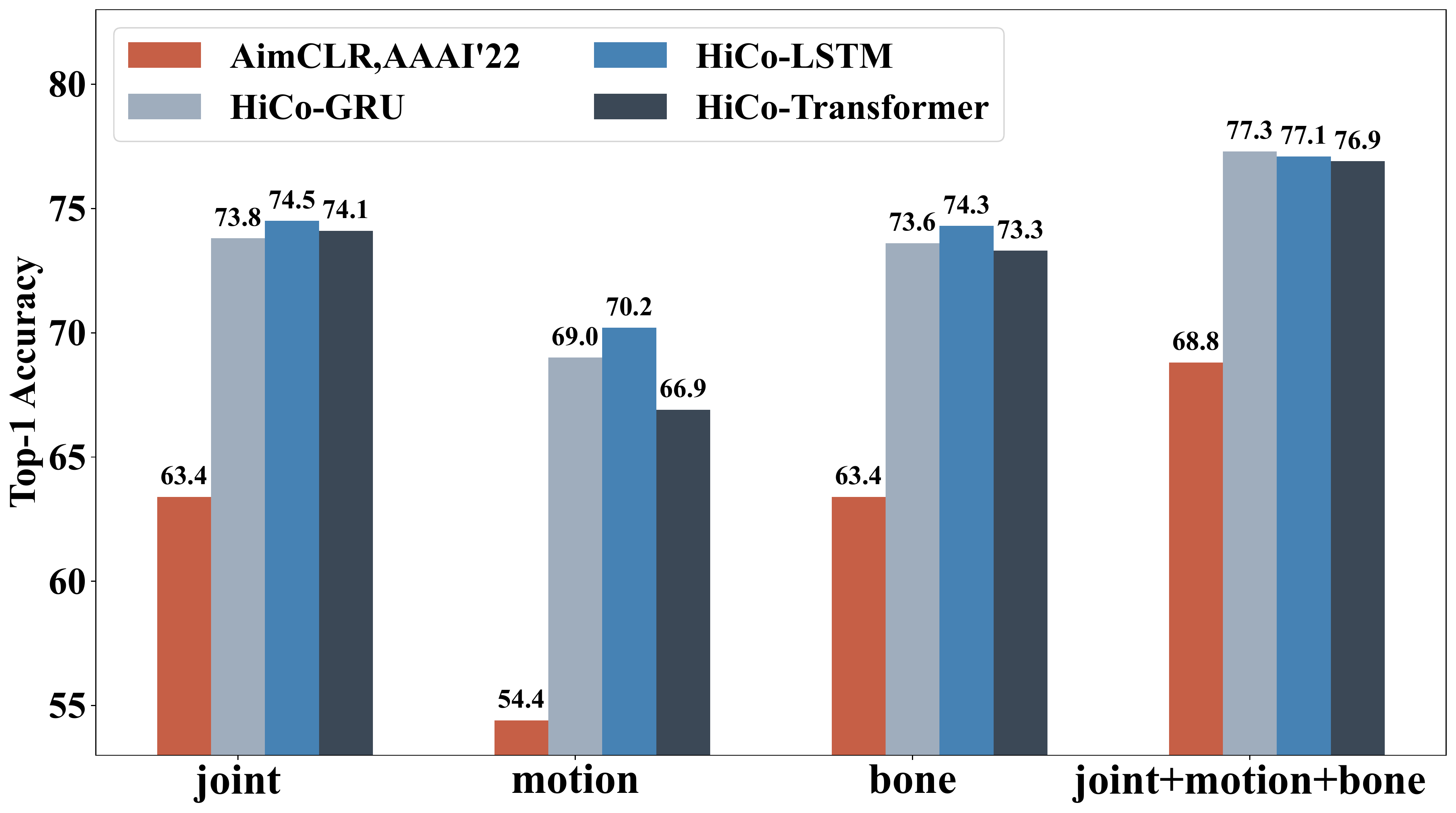}
\end{minipage}%
}
\subfigure[PKU-MMD I; x-sub]{
\begin{minipage}[t]{0.48\linewidth}
\centering
\includegraphics[width=3.2in]{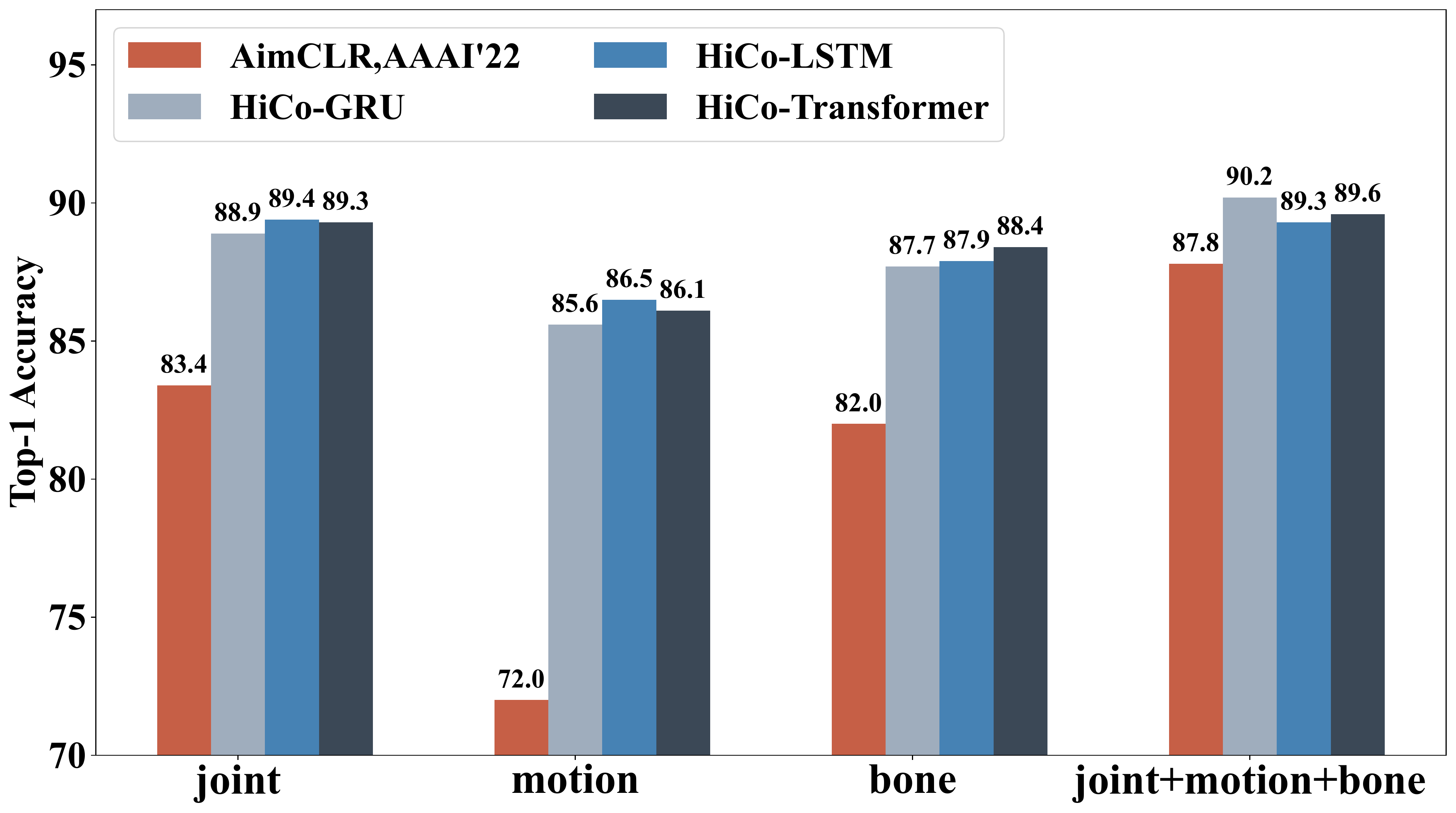}
\end{minipage}%
}
\centering
\caption{Performance comparison using different views of
skeleton sequences. Notice that CrosSCLR~\cite{li20213d} only reports their detailed performance on NTU-60, so we can not compare with it on the other datasets.
With the same views of human skeletons as input, our model leads by a large margin on all datasets.
}\label{fig:comparison_of_all_views}
\end{figure*}

\section{Additional Results}
\subsection{Comparison using various views on skeletons}\label{ssec:cmp}
In the main text we only report the corresponding results in terms of x-sub on NTU-60 due to the limited space. Here, we also the report results in terms of x-view on NTU-60 and the results on other datasets.
As shown in Figure \ref{fig:comparison_of_all_views}, using the same view of human skeletons as input, our model leads by a large margin on all datasets.

\begin{table*} [tb!]
\renewcommand{\arraystretch}{1.2}
\caption{
Complexity comparison with existing contrasitve learning based methods. 
All the results are evaluated in the same environment using one RTX 2080ti GPU.
}
\label{tab:cost}

\centering 
\scalebox{1.0}{
\begin{tabular}{@{}l*{5}c @{}}
\toprule
\textbf{Methods}& Memory (G) & Param (M)  & FLOPs (G)  \\ 
\hline
CrosSCLR \cite{li20213d}& 2.7  & 0.84 & 1.15  \\
ISC \cite{thoker2021skeleton}& 9.6  & 45.00 & 5.76 \\
AimCLR \cite{guo2022contrastive} & 4.0 & 0.84 & 1.15  \\
HiCo-GRU (ours) & 2.0 & 3.08 & 0.57 \\
HiCo-LSTM (ours) & 2.0 & 3.87 & 0.76 \\
HiCo-Transformer (ours) & 1.9  & 3.87 & 0.77\\
\bottomrule
\end{tabular}
 }
\end{table*}

\subsection{Complexity Comparison}\label{ssec:cost}
In this experiment, we conduct a complexity comparison of our proposed HiCo and other existing contrasitive learning.
Specifically, for each model, we report the GPU memory requirements using a mini-batch size of 64 during training, the number of parameters and FLOPs to encode a skeleton sequence during inference. Table \ref{tab:cost} summarizes the results.
Our HiCo variants require much less memory than ISC, and are comparable to CrosSCLR and AimCLR. 
Additionally, CrosSCLR and AimCLR have fewer parameters than ours but have more FLOPs. Although they use parameter-efficient ST-GCN as encoders, ST-GCN has high FLOPs as lots of convolution operations with shared parameters are employed in graphs. Besides, in our model FLOPs of UDM and Encoder are proportional to the sequence length. The sequence length is halved through UDM once, and the FLOPs are halved too. This mechanism makes our FLOPs not increase dramatically with the number of granularities. Note that the trainable parameters of encoders/UDMs in an individual branch are shared. All these lead to a small computation overhead.

\begin{figure*}[tb!] 
\centering
\subfigure[$L$=1; DBI: 6.60]{
\begin{minipage}[t]{0.23\linewidth}
\centering
\includegraphics[width=1.8in]{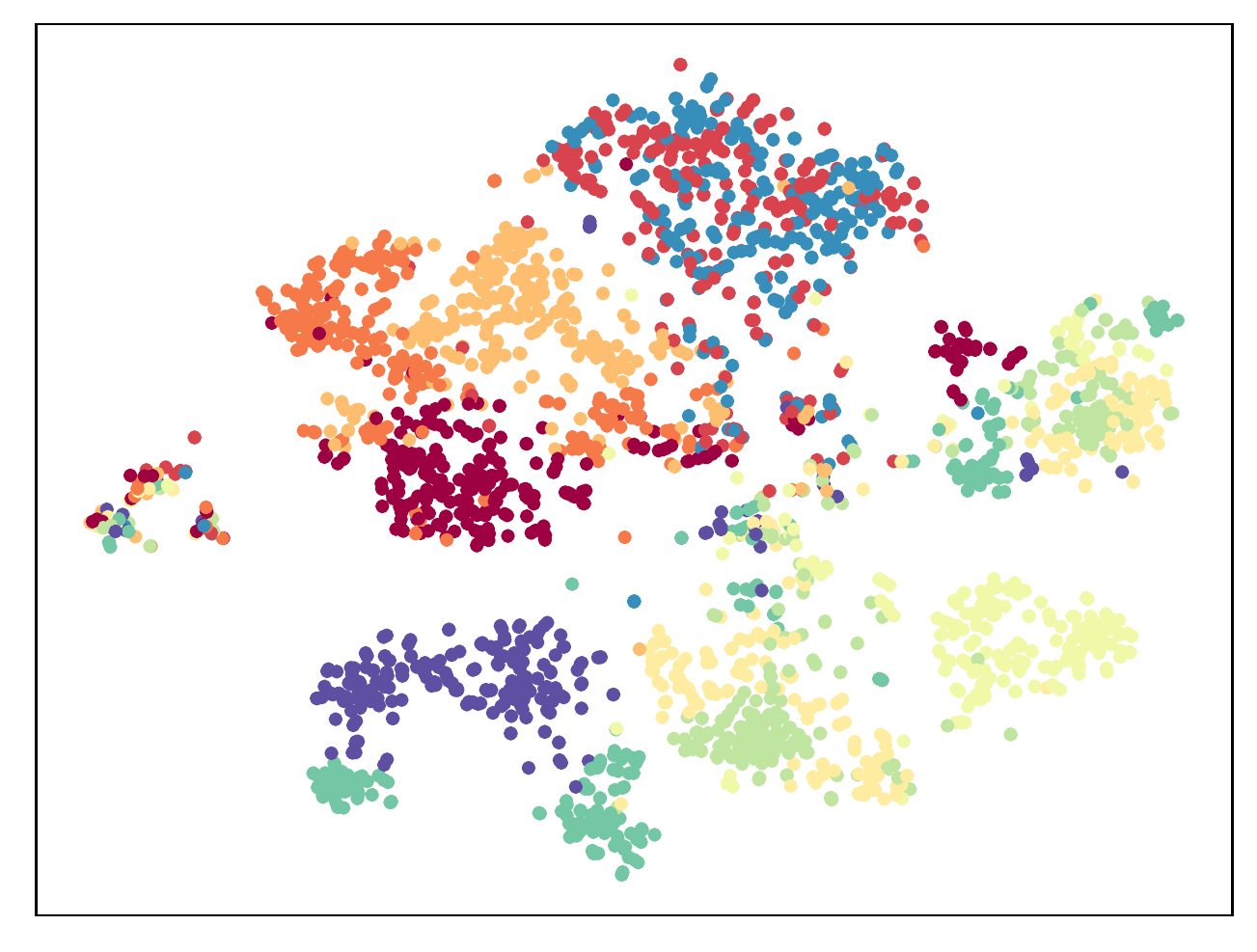}
\end{minipage}%
}
\subfigure[$L$=2; DBI: 6.29]{
\begin{minipage}[t]{0.23\linewidth}
\centering
\includegraphics[width=1.8in]{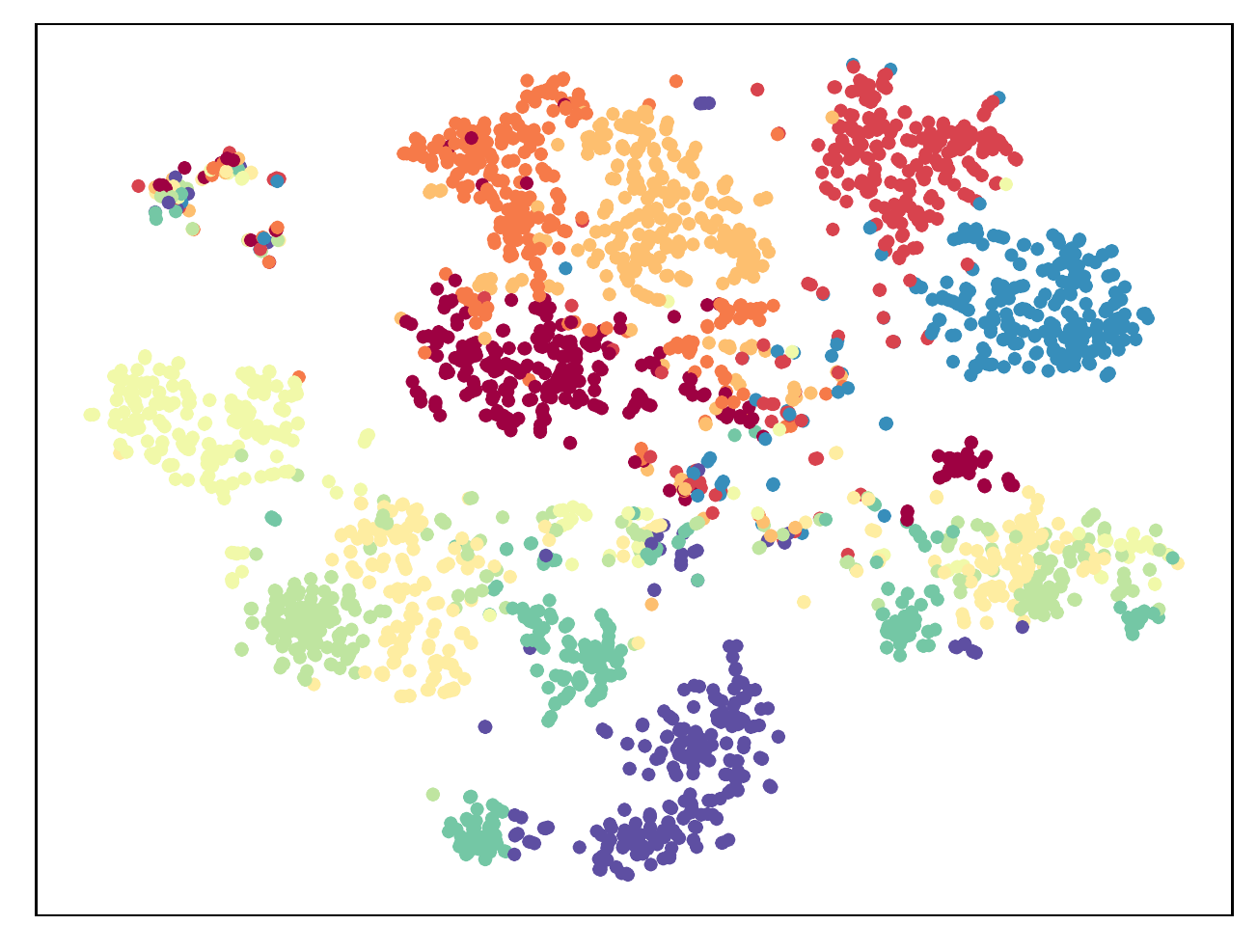}
\end{minipage}%
}
\subfigure[$L$=3; DBI: 6.13]{
\begin{minipage}[t]{0.23\linewidth}
\centering
\includegraphics[width=1.8in]{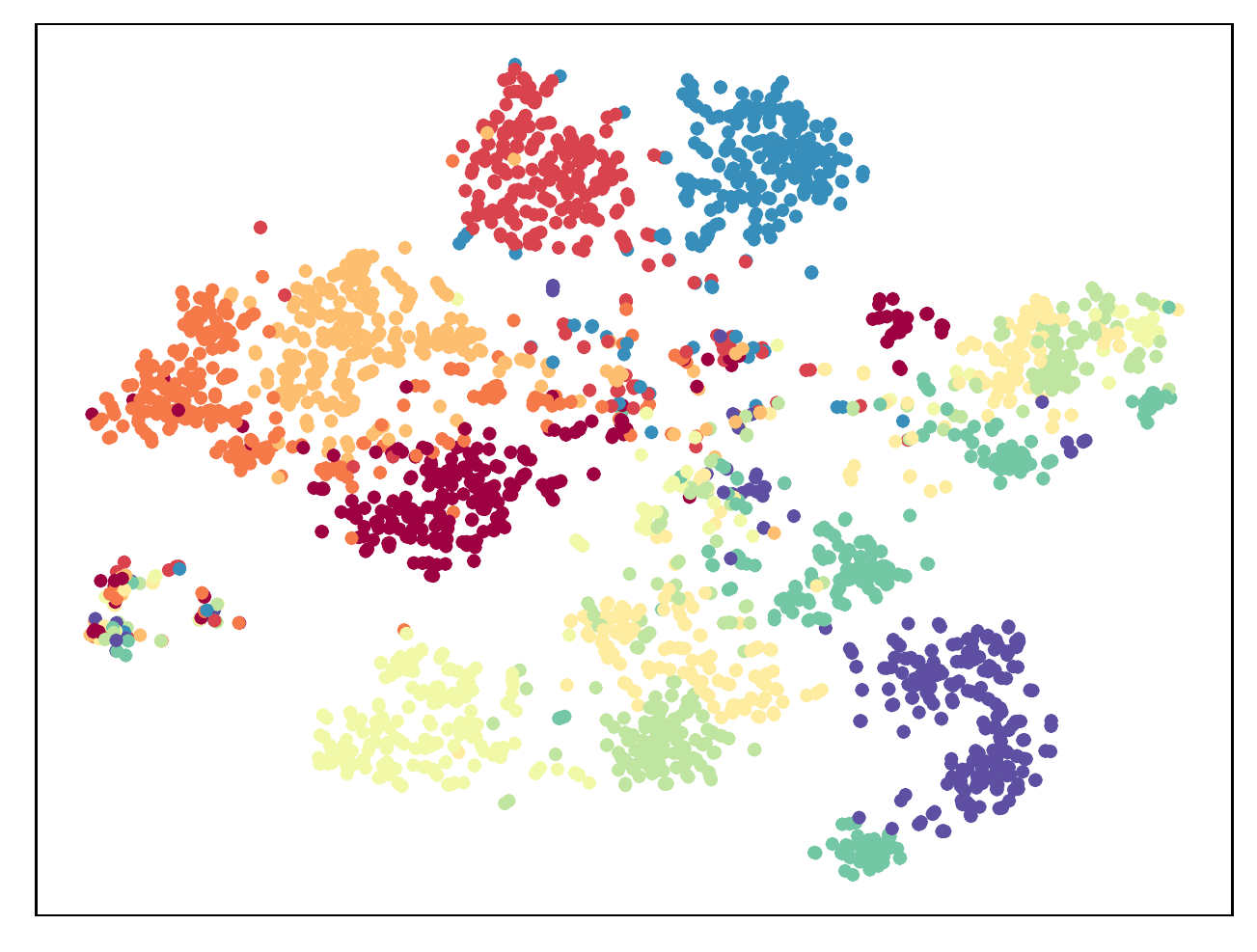}
\end{minipage}%
}
\subfigure[$L$=4; DBI: 5.56]{
\begin{minipage}[t]{0.23\linewidth}
\centering
\includegraphics[width=1.8in]{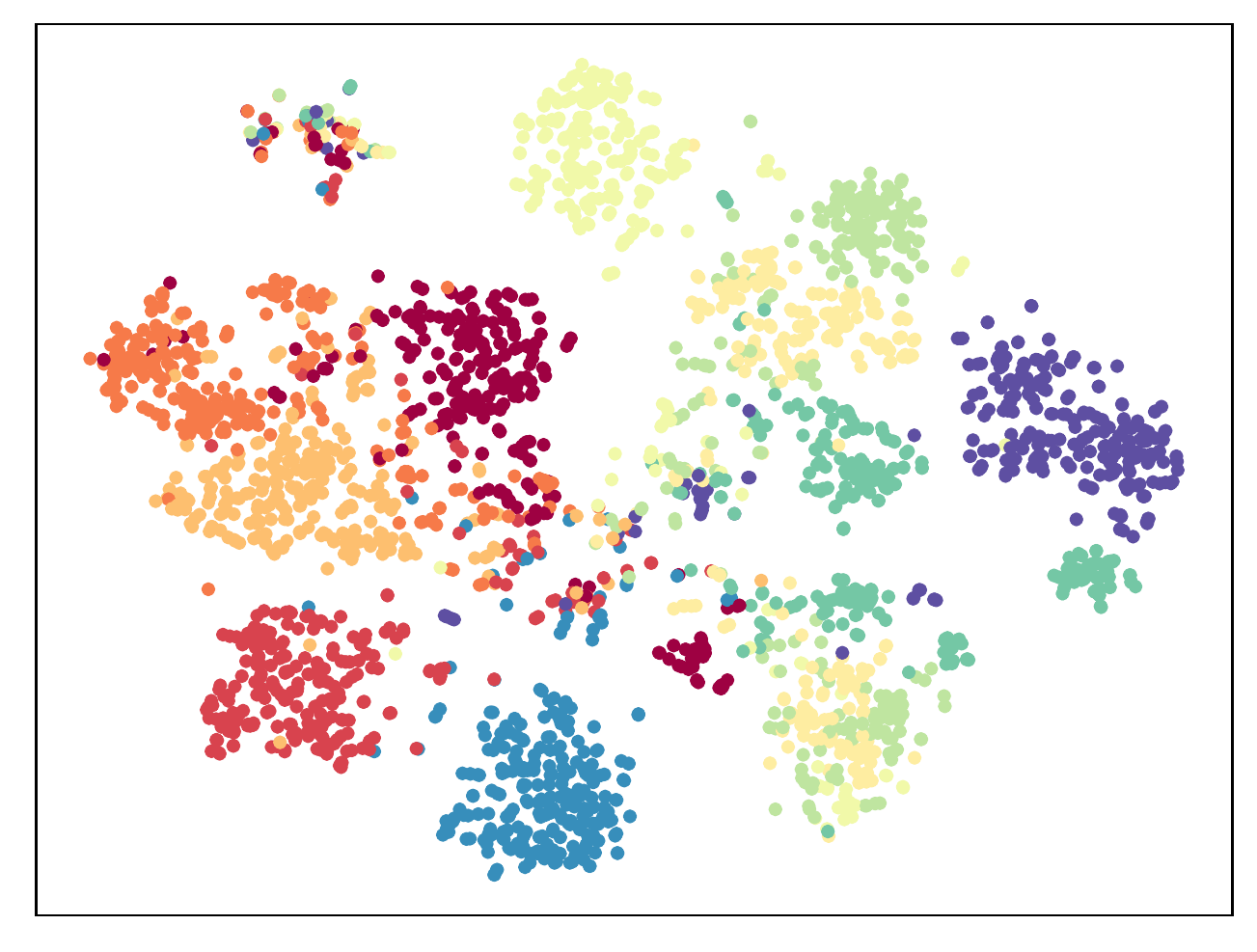}
\end{minipage}%
}

\centering
\caption{t-SNE visualization of the learned action representation obtained by our HiCo-Transformer with various settings on NTU-60. Dots with the same color indicate actions belonging to the same class. 
$L$ indites the number of granularities used in the models.
DBI denotes Davies–Bouldin index \cite{davies1979cluster} which is a common metric for evaluating clustering results. The lower DBI indicates the better learned representation.
More granularities used in HiCo-Transformer better representation learned generally, which further verifies the effectiveness of representing human skeletons into features of multiple granularities. 
}\label{fig:tsne}
\end{figure*}
\subsection{Visualization Results}\label{ssec:vr}
We utilize t-SNE \cite{van2008visualizing} to visualize the learned action representation obtained by our HiCo models with various settings, where all models utilize Transformer as encoders and are trained under the x-sub protocol. We randomly select 10 classes from the validation set for visualization.
Figure \ref{fig:tsne} shows the results of HiCo with a temporal branch using a different number of granularities.
Dots in Figure \ref{fig:tsne}(d) using four granularities are more clustered than that in the counterparts using less granularities (Figure \ref{fig:tsne}(a,b,c)).
It is consistent with the result of using more granularities achieves a lower DBI. 
The results again verify the effectiveness of representing human skeletons into features of multiple granularities.

\section{Implementation Details}\label{sec:id}
In this section, we describe the implementation details. Our source code will be released so readers of interest can have full access to every implementation detail.

\textit{Data Augmentation.} \label{ssec:da}
For a fair comparison, we directly use the data augmentation strategies used in \cite{thoker2021skeleton}, \ie shearing, joint jittering, and temporal cropping.
Specifically, shearing and joint jittering are linear transformations on the spatial dimension. The former randomly rotates the pose of skeletons and the latter randomly changes the position of the joint around the original location. Temporal cropping augments sequential data in temporal dimension by randomly selecting the starting frame within a certain range and resampling the sequence to a fixed length at different intervals.

\textit{Model Structure.} \label{ssec:ms}
We use two stacked GRU in a bidirectional manner as the S2S encoder of HiCo-GRU, and the LSTM counterpart for HiCo-LSTM. For HiCo-Transformer, we adopt a one-layer Transformer as the S2S encoder. It is worth noting that encoders for all granularities in the same branch are shared. The $C$ is set to 512, the output size of all encoders are set to 512, and the number of granularities $L$ is set to 4. The temporal max pooling and spatial max pooling are both 1D max pooling with kernel size of corresponding sequence length. The output dimension of projection MLP for feature projection is 128. 

\textit{Pre-training Details.} \label{ssec:pd}
During the training, we set the size of the dynamic dictionary queue to 2048, and the temperature value of $\tau$ is set to 0.2. For the optimizer, we adopt the stochastic gradient descent (SGD) method with a momentum of 0.9 and weight decay of 0.0001. The model is trained for 450 epochs with the initial learning rate of 0.01 and the learning rate is reduced by multiplying it by 0.1 at epoch 350, and the mini-batch size is set to 64.

\textit{Training Details for Downstream Tasks}.\label{ssec:dd}
For the skeleton-based action recognition task, an SGD optimizer with a momentum of 0.9 is adopted. We only train the linear classifier with the learning rate of 2, which is multiplied by 0.1 after the 50-th and 70-th epoch respectively. For transfer learning and semi-supervised learning, we finetune the encoder and classifier simultaneously, and settings are the same as that for skeleton-based action recognition, except the initial learning rate is set to 0.1.

\end{document}